\documentclass[10pt,twocolumn,letterpaper]{article}

\usepackage{iccv}

\usepackage{times}
\usepackage{graphicx}
\usepackage{epsfig}
\usepackage{svg}

\usepackage{amsmath}
\usepackage{amssymb}
\usepackage{booktabs}
\usepackage{epstopdf}
\usepackage{soul}
\usepackage{multirow}
\usepackage{comment}
\usepackage{latexcolors}
\usepackage{subcaption}
\usepackage{textcomp}
\usepackage{siunitx}
\usepackage{tabulary}
\usepackage{float}

\makeatletter\if@twocolumn\PassOptionsToPackage{switch}{lineno}\else\fi\makeatother
\usepackage[pagebackref,breaklinks,colorlinks]{hyperref}
\hypersetup{
    colorlinks=true,
    linkcolor=blue,
    urlcolor=magenta,
    citecolor=blue,
}

\usepackage{mathtools}
\usepackage{derivative}
\usepackage{nicefrac}
\usepackage{tikz}

\usepackage{soul}
\usepackage{multirow}
\usepackage{makecell}
\usepackage{comment}
\usepackage{latexcolors}
\usepackage{subcaption}
\usepackage{textcomp}
\usepackage{siunitx}
\usepackage{tabulary}
\usepackage{float}
\usepackage[boldmath]{numprint}
\definecolor{darkgreen}{rgb}{0.0,0.5,0}
\definecolor{darkblue}{rgb}{0.1,0.2,0.8}

\newcommand{\moniker}{DehazeNeRF}

\newcommand{\R}{\mathbb{R}}
\usepackage{paralist}  % tighter enumeration
\setdefaultleftmargin{2.5em}{2em}{1.7em}{1.5em}{1em}{1em}

\usepackage{xpunctuate}

\providecommand{\ie}[0]{i.e\xperiod}

\usepackage[capitalize]{cleveref}
\crefname{section}{Sec.}{Secs.}
\Crefname{section}{Section}{Sections}
\Crefname{table}{Table}{Tables}
\crefname{table}{Tab.}{Tabs.}
\Crefname{figure}{Figure}{Figures}
\crefname{figure}{Fig.}{Figs.}
\creflabelformat{equation}{#2#1#3}

\newcommand{\p}{\mathbf{p}}
\renewcommand{\r}{\mathbf{r}}
\newcommand{\x}{\mathbf{x}}
\newcommand{\y}{\mathbf{y}}
\renewcommand{\d}{\mathbf{d}}
\newcommand{\loss}{\mathcal{L}}
\newcommand{\mycolor}{pakistangreen}
\newcommand{\sdf}{{\color{\mycolor}f}}
\newcommand{\cnet}{{\color{\mycolor}c}}
\newcommand{\nn}[1]{{\color{\mycolor}#1}}

\newcolumntype{P}[1]{>{\centering\arraybackslash}p{#1}}
\newcolumntype{M}[1]{>{\centering\arraybackslash}m{#1}}

\iccvfinalcopy % *** Uncomment this line for the final submission
\def\iccvPaperID{6020} % *** Enter the ICCV Paper ID here

\ificcvfinal\fi

\begin{document}

%%%%%%%%% TITLE - PLEASE UPDATE
\title{\moniker{}: Multiple Image Haze Removal and 3D Shape Reconstruction using Neural Radiance Fields}

\author{
Wei-Ting Chen\textsuperscript{1,2*}\\
\and Wang Yifan\textsuperscript{1*}\\
\and Sy-Yen Kuo\textsuperscript{2} \\
\and Gordon Wetzstein\textsuperscript{1}\\
\and
\textsuperscript{1}Stanford University \hspace{1cm} \textsuperscript{2}National Taiwan University
}
\maketitle

%%%%%%%%% ABSTRACT
\begin{abstract}
Neural radiance fields (NeRFs) have demonstrated state-of-the-art performance for 3D computer vision tasks, including novel view synthesis and 3D shape reconstruction. Yet, these methods fail in adverse weather conditions that are crucial for applications such as autonomous driving. To address this challenge, we introduce DehazeNeRF as a framework that robustly operates in hazy conditions. \moniker{} extends the volume rendering equation by physically realistic terms that model atmospheric scattering. These act as inductive network biases in our pipeline and, together with several regularization strategies, allow DehazeNeRF to demonstrate successful multi-view haze removal, novel view synthesis, and 3D shape reconstruction where existing approaches fail.
\end{abstract}

\let\thefootnote\relax\footnotetext{* indicates equal contribution.}
\section{Introduction}
Neural radiance fields (NeRFs)~\cite{mildenhall2020nerf} have emerged as a powerful approach to solving 3D computer vision problems, such as novel view synthesis and 3D reconstruction~\cite{tewari2022advances}. Existing NeRF\footnote{We refer to neural radiance fields and neural surfaces~\cite{wang2021neus} collectively as ``NeRF'' in this paper.} approaches are successful in many scenarios, but they fail to accurately reconstruct the geometry and appearance of a 3D scene in adverse weather conditions such as haze (see \cref{fig:real_qualitative}). This poses a critical limiting factor for many real-life applications, including autonomous driving among others.

The problem of applying NeRF to hazy input images is the ambiguity between light-reflecting solid surfaces and light-scattering atmospheric particles in free space -- both are interchangeably modeled as a light-emitting volume by NeRF and cannot be disambiguated.
Most state-of-the-art dehazing methods adopt feedforward neural networks~\cite{singh2019comprehensive,ancuti2019ntire}, but applying these to the input views as a preprocessing step to standard NeRFs leads to poor reconstruction results, in part because these techniques tend to overfit to the training data and generalize poorly to real-world data~\cite{Qu_2019_CVPR,shao2020domain}.

\begin{figure}[t!]
\centering \includegraphics[width=0.8\linewidth]{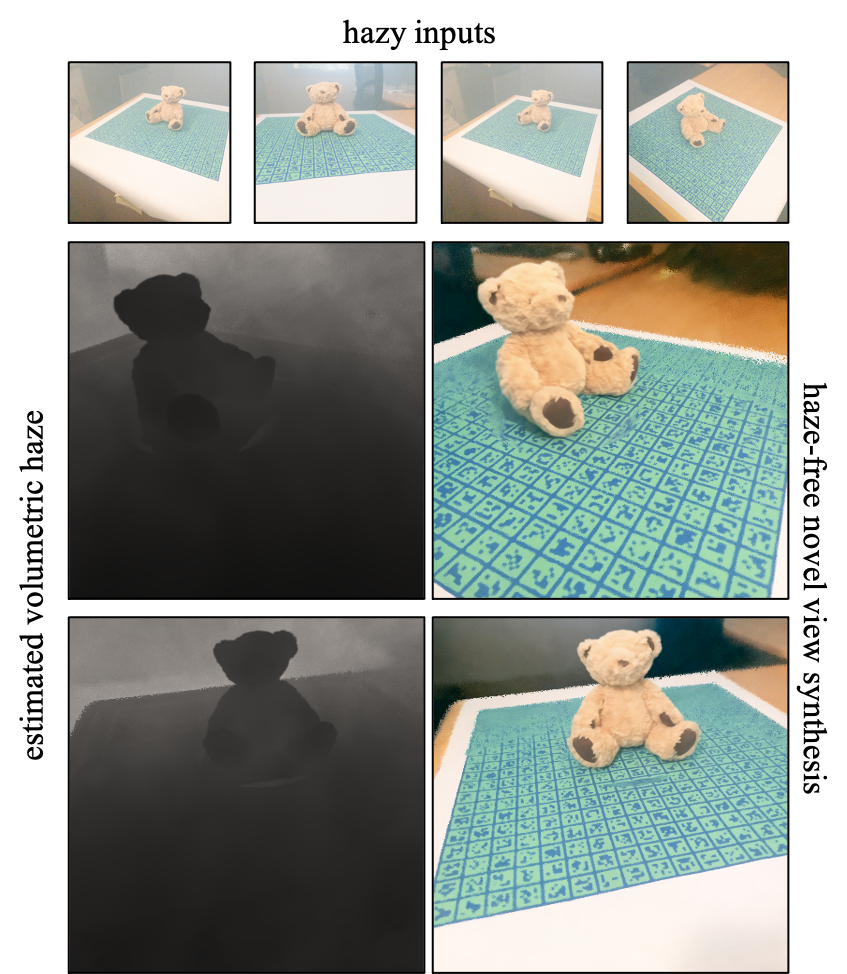}
\makeatother
    \caption{Given a set of posed hazy images, our method estimates the haze-free scene using a neural radiance field that includes physically based terms for scattering. Here, we show results of an experimentally captured scene.}
	\vspace{-0.5cm}\label{fig:teaser_figure}
\end{figure}
To address this limitation, we develop a differentiable physically based 3D haze model that seamlessly integrates into the volume rendering framework used by NeRF.
In the absence of scattering, our model is equivalent to NeRF. %and, for single image input, it adequately models the image formation used by existing single image dehazing methods.
For the inverse problem of estimating the scene parameters from multiple posed input views, our 3D haze model is used in conjunction with several regularization strategies that disentangle solid object surfaces from scattering volumetric particles.
Our proposed framework jointly learns the 3D shape and haze components of a scene and allows for significantly improved and multi-view-consistent novel view synthesis as well as more accurate 3D shape reconstruction compared to previous work.
Our approach does not require large collections of paired hazy-clear images and it circumvents the sim2real gap of data-driven dehazing methods by building an inductive bias based on the physics of scattering into the network architecture.

Specifically, our contributions include:
\begin{compactitem}
\item extending the volume rendering equation of NeRF by a physically based 3D haze image formation model to accurately model the in-scattering phenomenon prevalent in hazy conditions;
\item introducing multiple physically inspired inductive biases as well as optimization regularizers to effectively disambiguate the surface appearance thus achieving accurate clear-view appearance and geometry reconstruction using only hazy image as inputs.
\end{compactitem}
We demonstrate state-of-the-art results using simulation and experimental data for 3D multi-view construction and novel view synthesis under hazy condition.
Code and data will be made available \footnote{\url{https://www.computationalimaging.org/publications/dehazenerf}} for research purposes.

\section{Related Work}\label{sec:relatedwork}
\moniker{} integrates knowledge from several areas of research.
We briefly review these next.

\subsection{3D Scene Reconstruction with NeRF}
Neural radiance fields (NeRFs)~\cite{mildenhall2020nerf} were introduced for novel view synthesis applications.
By combining differentiable volumetric rendering from classic computer graphics and learnable radiance field, NeRF can jointly optimize for the geometry and appearance of a static 3D scene from posed RGB images.
As the performance of the original NeRF noticeably deteriorates under imperfect capturing conditions, much research has investigated the robustness of NeRF under a variety of challenging conditions, such as blur~\cite{ma2022deblur}, noise~\cite{pearl2022nan}, reflection~\cite{guo2022nerfren}, low resolution~\cite{wang2022nerf}, low dynamic range~\cite{huang2022hdr,mildenhall2022nerf}, and occlusion~\cite{martin2021nerf,chen2022hallucinated}.
In this paper, we address haze, an unexplored yet common scenario in real-life captures.
Our method leverages a scattering-aware rendering equation to model the haze phenomenon and disambiguate the surface and haze by using physically inspired inductive biases.
Among those, we adopt a surface-like density field parameterization~\cite{wang2021neus} to encourage solid geometry, although other surface priors may be equally applied in our method~\cite{niemeyer2022regnerf,yariv2021volume,wang2021neus, oechsle2021unisurf}.

\subsection{Dehazing}
We provide a brief summary of existing dehazing methods and refer to~\cite{singh2019comprehensive} for a comprehensive review.
Early works aim to design strong priors to recover the transmission map, global atmospheric light, and scene radiance~\cite{zhu2015fast,bui2017single,fattal2014dehazing, he2010single, berman2016non}, thus they struggle on scenes where these priors do not hold~\cite{liu2019learning}.
More recently, data-driven methods~\cite{liu2019learning,chen2019pms,dong2020multi,mehri2021mprnet,zamir2021multi,guo2022image} achieve better quality using deep learning and large datasets~\cite{li2018benchmarking}, but have difficulty generalizing to images in the real world~\cite{shao2020domain}.
More related to our work are multi-image dehazing methods, which exploit the consistency between neighboring frames.
Similar to single-image approaches, multi-image approaches can be divided into prior-based or data-driven categories.
The former adds photo-consistency regularization between neighboring frames using jointly estimated or known depth~\cite{zhang2011video,li2015simultaneous,fujimura2020dehazing}, while the latter achieves more temporal stability by fusing features from a small number of neighboring frames~\cite{ren2018deep,wang2019edvr,zhang2021learning}.
Yet both approaches are subject to the aforementioned problems exhibited in the single-image scenario.
Furthermore, none of these dehazing methods estimates a holistic 3D structure, and therefore they are not directly applicable to novel view synthesis.

\paragraph{Estimating Scattering Coefficients.}
In addition to estimating the 3D geometry and haze-free appearance in novel views, we also jointly optimize for the scattering coefficient.
Prior work mainly computes the scattering coefficient based on handcrafted priors such as average saturation~\cite{gu2017single}, dark channel mean~\cite{chung2022image}, polarization~\cite{schechner2001instant}, and gamma correction~\cite{ju2019idgcp}.
In data-driven approaches, the scattering coefficient can be estimated explicitly as an intermediary output~\cite{yang2022self,wang2021fully}.
Our method directly estimates the scattering coefficient in 3D, while at the same time utilizing 2D priors such as dark channel mean to address the ambiguity with other physical properties such as airlight.

\paragraph{Seeing through Scattering Media.}
Various methods have been developed to solve the challenging problem of imaging through and within scattering media. These methods can be classified as interference of light, relying on ballistic photons, and being based on diffuse optical tomography. For methods based on interference of light, they leverage information in the speckle pattern to reconstruct the image~\cite{popoff2010image,katz2014non,bertolotti2012non} or adopt wavefront shaping to focus light through or within scattering media~\cite{horstmeyer2015guidestar,vellekoop2007focusing}. On the other hand, ballistic photons can avoid scattered photons since they can travel through a medium without scattering and can be isolated by adopting time-gating~\cite{redo2016terahertz,wang1991ballistic}, coherence-gating~\cite{indebetouw2000imaging,dunsby2003techniques}, or coherent probing and detection of a target at different illumination angles~\cite{kang2015imaging}. The last class is to adopt non-line-of-sight imaging techniques~\cite{velten2012recovering,liu2019non,o2018confocal,lindell2019wave,faccio2020non,liu2020phasor,young2020non} or diffuse optical tomography~\cite{boas2001imaging,gibson2009diffuse,Lindell:2020:CDT} to recover objects by modeling and inverting scattering of light explicitly.
However, these methods all require exotic hardware setups, which are usually expensive, and are limited to some scenarios such as long propagation distance of light or scale of resolution.
Our approach removes the scattering effect using inverse rendering, requiring only a set of RGB images of a hazy scene.
Similar principles have been explored in concurrent and independent work~\cite{levy2023seathrunerf} for underwater scene reconstruction.
\section{Method}
Our method, {\moniker}, extends the volume rendering equation to accurately reconstruct the geometry and appearance robust to hazy conditions.
Our key idea is to introduce a series of important biases in the network architecture along with regularizers in the loss function that together underpin physically based scattering phenomena.

\subsection{Preliminary on Neural Radiance Fields}\label{sec:nerf}
Neural Radiance Fields (NeRFs)~\cite{mildenhall2020nerf} map a 3D sample point \(\p\) into a color $\mathbf{c}$ and volume density $\sigma$.
Considering only emission from classic volume rendering~\cite{kajiya1984ray,tagliasacchi2022volume}, the expected color ${C}(\r)$ of a camera ray $\r(t)=\mathbf{o} + t\mathbf{d}$ with the near and far boundary $t_n$ and $t_f$ can be written as
\begin{gather}
	{C}(\r, \mathbf{d})=\int_{t_n}^{t_f}T(t)\sigma(\r(t))c(\r(t), \mathbf{d}) \ dt \;\textrm{with} \label{eq:nerf}\\
    T(t)=\mathrm{exp}\left( - \int_{t_n}^{t}\sigma(\r(t')) \ dt'\right),
	\label{eq:occlusion}
\end{gather}
where \(T(t)\) is the accumulated transmittance between the ray section \(t_{n}\) to \(t \).
The predicted pixel value is then compared to the ground truth $\widehat{C}(\r,\d)$ for optimization.

\subsection{3D Haze Formation}\label{sec:rte_haze}
To address the 3D dehazing problem, we propose an alternative rendering equation to the image formation model.
We start from the radiative transfer equation (RTE)~\cite{chandrasekhar2013radiative,van1999multiple}, which describes the behavior of light in a medium that absorbs, scatters and emits radiation.
Assuming, a ray \(\r\left( t \right) = \mathbf{o} + t\d\) hits a surface point at \(\r\left( t_{0} \right)\), the incident radiance at the near image plane \(t_{n}\) can be divided into three parts~\cite{pharr2016physically}:
{\small
\begin{align}
C(\r, \d) &= C_{\textrm{emission}}(\r) + C_{\textrm{surface}}(\r) + C_{\textrm{in-scattering}}(\r)\nonumber\\
C_{\textrm{emission}}(\r, \d) &=
\int_{t_{n}}^{t_{0}}\epsilon\left(\r\left( t\right),\d\right)T_{\sigma_{t}}\left( t\right)dt\nonumber\\
C_{\textrm{surface}}(\r, \d) & =C_e\left(\r\left( t_{0} \right),\d\right)T_{\sigma_{t}}\left( t_{0}\right)\nonumber\\
C_{\textrm{in-scattering}}(\r, \d) &=
\int_{t_{n}}^{t_{0}}c_{\textrm{s}}\left( \r\left( t \right), \d \right)\sigma_{s}\left(\r\left( t \right)\right)T_{\sigma_{t}}\left( t \right)dt,\nonumber
\end{align}
}
where \(\epsilon\) is the emission, \(C_{e}\) is the outgoing radiance at the surface intersection, \(c_{\textrm{s}}\left(\r\left( t \right), \d \right)\) is the in-scattered light and \(\sigma_{s}\) is the scattering coefficient.
In particular, the transmittance here is computed from the attenuation coefficient \(\sigma_{t}\), \ie,
\(T_{\sigma_{t}}\left( t\right)=\exp\left( -\int_{t_{n}}^{t}\sigma_{t}(t')dt' \right)\),
where \(\sigma_{t}=\sigma_{a} + \sigma_{s}\) including the absorption and out-scattering effect.
For common haze formation, the participating particles are considered non-luminous~\cite{narasimhan2003contrast}, therefore we can drop the emission part, which leads to
{
\small
\begin{align}
\begin{split}
C(\r,\d)= {} & C_e(\r\left( t_{0} \right),\d)T_{\sigma_{t}}\left( t_{0} \right)+\\
&\int_{t_{n}}^{t_{0}}c_{\textrm{s}}\left( \r\left( t \right), \d \right)\sigma_{s}\left(\r\left( t \right)\right)T_{\sigma_{t}}\left( t \right)dt.
\end{split}
\label{eq:RTE_Haze}
\end{align}
}

Following NeRF~\cite{mildenhall2020nerf}, we represent the surface as a continuous density field with emission \(\epsilon\left(\r\left(t\right), \d\right)\coloneqq c\left(\r\left( t \right),\d\right)\sigma\left(\r\left( t \right)\right)\).
Meanwhile, the absorption part in the attenuation \(\sigma_{t}\) can be interpreted as the surface density \(\sigma\), since the volume density $\sigma$ is equal to absorption coefficient $\sigma_{a}$ in that they both determine the probability of a photon or a ray terminating at a given location.
As a result, we can write the rendering equation as
{\small
\begin{align}\begin{split}
    C(\r,\d)=&
    \underbrace{\int_{t_{n}}^{t_{0}}c(\r(t),\d)\sigma(t)T_{\sigma+\sigma_{s}}\left( t \right)dt}_{C_{\textrm{Surface}}} +\\
    &\underbrace{\int_{t_{n}}^{t_{0}}c_{s}(\r(t))\sigma_{s}(t)T_{\sigma+\sigma_{s}}\left( t \right)dt}_{C_{\textrm{Haze}}}.
    \label{eq:3D_haze_formation}
\end{split}
\end{align}
}
\cref{eq:3D_haze_formation} formally disentangles the surface and haze, represented by \(\left\{ c, \sigma \right\}\) and \(\left\{  c_{s}, \sigma_{s} \right\}\) respectively, in a principled manner.
Once successfully optimized (see the next Section), the clear-view surfaces can be recovered using \(\left\{ c, \sigma \right\}\):
\begin{equation}
C(\r,\d)=
\int_{t_{n}}^{t_{0}}c(\r(t),\d)\sigma(t)T_{\sigma}\left( t \right)dt\label{eq:clear_view}.
\end{equation}

\begin{figure}[t!]
\centering \includegraphics[width=\linewidth]{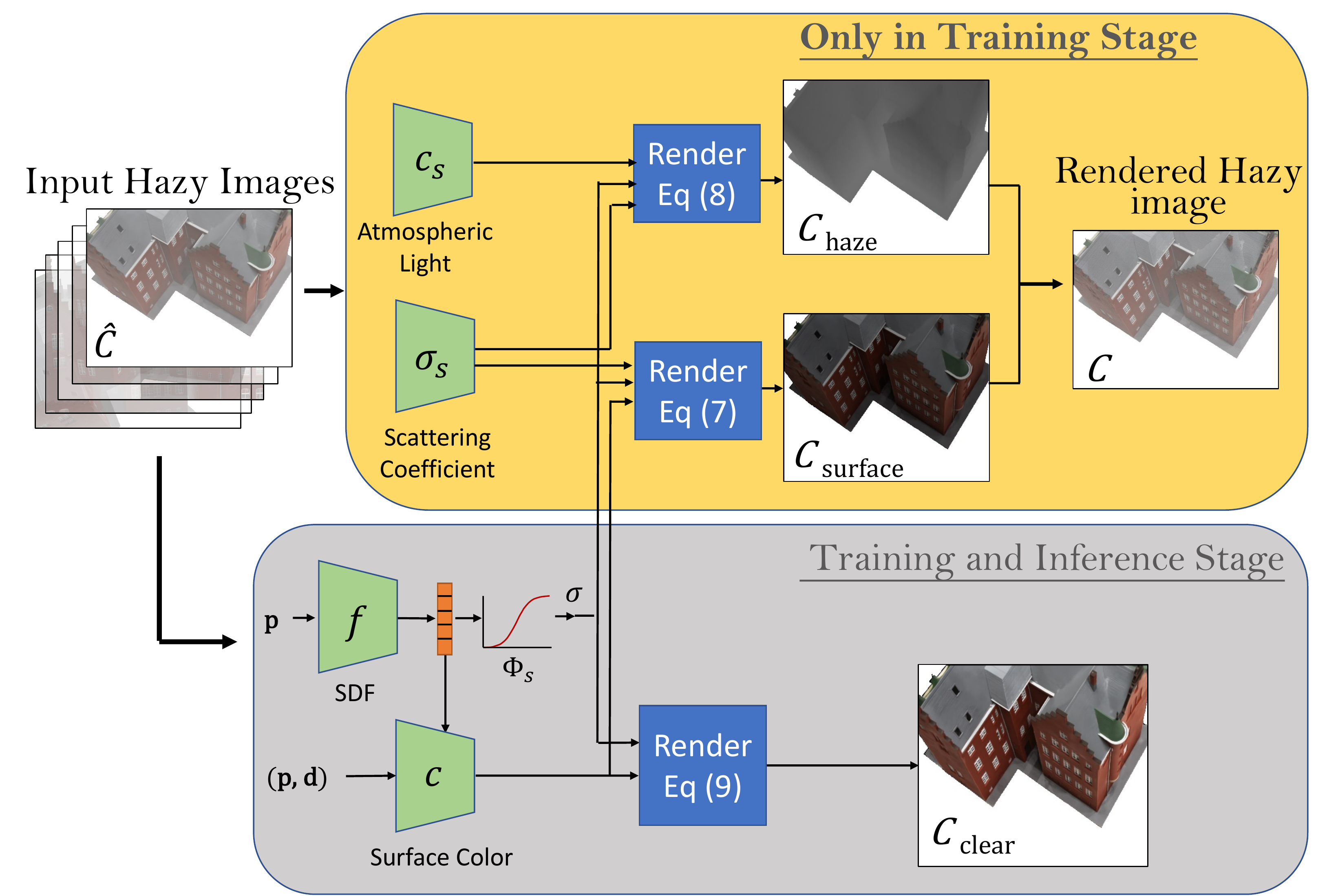}
\makeatother
\caption{\textbf{\moniker{} architecture.} Given a set of hazy images, our method augments the existing NeRF pipeline (gray) with a haze module (yellow), which explicitly models the scattering phenomenon using atmospheric light and scattering coefficient. During training, we render the hazy reconstruction as a composition of surface and haze, which is compared to the input hazy images to optimize the learnable parameters (in green) jointly. During inference, we use the surface module (gray) to render clear views.}
\vspace{-0.5cm}\label{fig:architecture}
\end{figure}

\subsection{Haze-aware Neural Radiance Field}\label{sec:dehaze_nerf}
Given multiple images of a hazy scene, we aim to jointly optimize for the surface appearance and geometry, \(\left\{ c, \sigma \right\}\) as well as the haze's scattering coefficient and in-scattered light (atmospheric light),  \(\left\{c_{s}, \sigma_{s} \right\}\) based on the enhanced scattering-aware rendering equation~\cref{eq:3D_haze_formation}.
However, the effects of these variables are interdependent. In order to correctly disentangle them, our model adopts suitable architecture designs and training regularizers to capture the distinct physical properties of haze and surface.
An overview of \moniker{} is illustrated in \cref{fig:architecture}.

\paragraph{Architecture.} Now we introduce inductive biases to match the physical properties of haze and surface.
For clarity, we highlight the quantities directly modeled by neural networks in \nn{green}.

\emph{Modeling a Surface.} Recall our goal is to learn the surface appearance and geometry, \(\left\{ c, \sigma \right\}\).
Similar to previous works~\cite{mildenhall2020nerf}, we model the appearance \(\cnet\left( \p, \d \right)\) with an MLP, which takes the sample location \(\p\) and viewing direction \(\d\) as inputs.
However, in order to encourage volume density \(\sigma\) to form a well-defined solid surface, instead of directly learning the volume density, we adopt the reparameterization of the volume density using signed distance function (SDF), \(\sdf\left( \r\left( t \right) \right)\in \R\), as proposed in NeuS~\cite{wang2021neus,wang2022hfs}.
The modified surface volume density \(\sigma\left( \r\left( t \right) \right) \), referred to as opaque density, can be parameterized as \(\sdf\left( \r\left( t \right) \right)\):
\begin{equation}
\sigma\left( \r\left( t \right) \right) = s\left( \Phi_{s}\left( \sdf\left( \r\left( t \right) \right)\right) -1 \right)\nabla \sdf\left( \r\left( t \right) \right)\mathbf{d},\label{eq:hfneus-sigma}
\end{equation}
where $\Phi_{s}(x)$ is the sigmoid function $\Phi_s(x) = (1 + e^{-sx})^{-1}$, whose derivative is a bell-shaped density function centered at 0 and has a learnable standard deviation of \(\nicefrac{1}{s}\).
We derive the discrete approximate following~\cite{mildenhall2020nerf,tagliasacchi2022volume}.
It samples $n$ points $\left\{ \p_{i}=\mathbf{o}+t_n\mathbf{d}|n=1,...,N,t_n<t_{n+1} \right\}$ along the ray.
The approximate pixel color of the ray is computed based on quadrature rule~\cite{max1995optical}, yielding
\begin{align}\begin{gathered}
C_{\textrm{surface}}(\r,\d) = \sum_{n=1}^{N}\frac{\sigma^{n}}{\sigma_{t}^{n}} T_{t}^{n}\alpha_{t}^{n}\nn{c}^{n} \textrm{ with } T_{t}^{n}=\prod_{m=1}^{n-1}\left(1 - \alpha_{t}^{m}\right) \label{eq:C_surface},
\end{gathered}\end{align}
where \(\alpha_{t}\) denotes the discrete \(\alpha\)-compositional weight defined as~\cite{wang2021neus,wang2022hfs}
\begin{equation}
 \resizebox{1\hsize}{!}{
 $
    \alpha_{t}^{n}=\textsc{clamp}\left( 1-\exp\left( -\sigma_{t}^{n}\delta^{n} \right),0, 1 \right) \textrm{ with } \delta^{n}=t^{n+1}-t^{n}\label{eq:alpha}\nonumber,$}
\end{equation}
where \(\sigma_{t}^{n}=\sigma^{n}+\nn{\sigma_{s}}^{n}\) denotes the total attenuation at sample \(n\), including the attenuation due to surface occlusion and the out-scattering.

\emph{Modeling Haze.} We use a low-frequency prior to compute the scattering coefficient and atmospheric light, \(\left\{c_{s}, \sigma_{s} \right\}\), since these components usually vary slowly in a common hazy scenes~\cite{li2015simultaneous}.
In practice, we use a small band-limited \textsc{MLP}~\cite{lindell2022bacon} for the scattering coefficient \(\sigma_{s}\) to capture inhomogenous haze.
Analogous to \cref{eq:C_surface}, the haze color can be approximated as
% \begin{equation}
% \begin{gathered}
% C_{\textrm{haze}}(\r) = \sum_{i=1}^{n}\frac{\nn{\sigma_{s}}^{n}}{\sigma_{t}^{n}} T_{t}^{n}\alpha_{t}^{n}\nn{c_{s}}^{n}.\label{eq:C_haze}
% \end{gathered}
% \end{equation}
\begin{equation}
\begin{gathered}
C_{\textrm{haze}}(\r) = \sum_{n=1}^{N}\frac{\nn{\sigma_{s}}^{n}}{\sigma_{t}^{n}} T_{t}^{n}\alpha_{t}^{n}\nn{c_{s}}^{n}.\label{eq:C_haze}
\end{gathered}
\end{equation}
During optimization, the color for an arbitrary input hazy image can be written as $C = C_{\textrm{surface}} + C_{\textrm{haze}}$.
At test time, we can reconstruct the clear-view color by discretizing \cref{eq:clear_view}, namely:
\begin{gather}
 C_{\textrm{clear}}\left( \r,\d \right) = \sum_{n=1}^{N}T_{\sigma}^{n}\alpha^{n} \nn{c}^{n}, \label{eq:clear_view_discrete}\\
 \resizebox{1\hsize}{!}{
 $T_{\sigma}^{n} = \prod_{j=1}^{n-1}\left(1 - \alpha^{j}\right)\, \textrm{and }\, \alpha^{n} = \textsc{clamp}\left( 1 - \exp\left( -\sigma^{n}\delta^{n} \right),0, 1 \right).\nonumber$}
\end{gather}
\paragraph{Optimization.} While the inductive biases separate the high-frequency surface appearance and geometry from the low-frequency color and density of the scattering medium, we introduce further regularizers to guide the optimization process to converge to more plausible clear-view geometry and color.

\emph{Koschmieder Consistency.}
Given an accurate depth map \(D\), assuming globally constant scattering coefficient \(\bar{\sigma}_{s}\) and airlight \(\bar{c}_{s}\), the relation between a clear-view image \(C_{\textrm{clear}}\) and the hazy image \(C\) can be described by the Koschmieder law~\cite{israel1959koschmieders} as
\begin{equation}
\resizebox{0.88\hsize}{!}{
\(C(\r)=C_{\textrm{clear}}(\r)\exp(-\bar{\sigma}_{s} D(\r))+\bar{c}_{s}(1-\exp(-\bar{\sigma}_{s} D(\r)))\).
}\label{eq:koschmieder}
\end{equation}
This model is widely adopted as the basis for image-based single and multiview dehazing.
The Koschmider model is an approximation of our rendering equation~\cref{eq:3D_haze_formation} under the assumption of
spatially-invariant (i.e., homogeneous) scattering coefficient and an ideal surface
\begin{align}
C_{\textrm{surface}}\left( \r \right) & \approx C_{\textrm{clear}}(\r)\exp(-\bar{\sigma}_{s} D(\r)) = \tilde{C}_{\textrm{surface}}\left( \r \right)\\
C_{\textrm{haze}}\left( \r \right) & \approx \bar{c}_{s}(1-\exp(-\bar{\sigma}_{s} D(\r)) = \tilde{C}_{\textrm{haze}}\left( \r \right),
\end{align}

We promote this relation with
\begin{align}
&\loss_{\textrm{2D}} = \left\|C_{\textrm{surface}}\left( \r \right) -  \tilde{C}_{\textrm{surface}}\left( \r \right)\right\|_{1} \\+
&\left\| C_{\textrm{haze}}\left( \r \right)\! - \!\tilde{C}_{\textrm{haze}}\left( \r \right)\right\|_{1} \!\!+\!
 \left\| C\! -\! \tilde{C}_{\textrm{surface}}\left( \r \right)\! -\! \tilde{C}_{\textrm{haze}}\left( \r \right)\right\|_{1}\!, \nonumber
\end{align}
where \(\bar{\sigma}_{s}\) and \(\bar{c}_{s}\) are the average over the samples on the ray, while
the depth value \(D\left( \r \right)\) is computed via the learned surface geometry~\cite{mildenhall2020nerf,yu2022monosdf} by accumulating over ray-length over all the samples on a ray:
\begin{equation}
    D\left( \r \right) = \sum_{n=1}^{N} T_{\sigma}^{n}\alpha^{n}t^{n}.
\end{equation}

\emph{Color Prior.}
Without knowing the original image, the heavily attenuated color in the hazy image can be explained by the haze but also by a dull surface color.
In order to reconstruct plausible clear-view colors, we adopt the popular 2D prior widely used in image-based dehazing methods, Dark Channel Prior (DCP)~\cite{he2010single}, which arises from the observation, that for most pixels in a natural haze-free image, the minimum of three color channels is close to zero.
We apply this prior to the estimated clear image \(C_{\textrm{clear}}\)
\begin{align}
DC(C_{\textrm{clear}})\left(\x\right)&=\underset{\y\in\Omega\left(\x\right)}{\min}\left(\underset{c\in\left\{r,g,b\right\}}{\min}C_\textrm{clear}^{c}\left(\y\right)\right),
\label{eq:DCP_definition}\\
\loss_{\textrm{dcp}}&=\frac{1}{K}\sum\limits_{k=1}^{K}\Vert DC\left(C_{\textrm{clear}}\right)\Vert_{1}.
\label{eq:loss_dcp}
\end{align}

\subsection{Implementation Details}
We adopt the same setting as that in HF-NeuS~\cite{wang2021neus} wherever possible.
This includes the MLPs for the surface SDF, \(\sdf\) and the view-dependent surface color, \(\cnet\), as well as the sampling strategy, the background composition, and learning rate schedule.

\paragraph{Loss.}
Our loss is composed of several terms:
\begin{equation}
    \loss = \loss_{\textrm{color}} + \lambda\loss_{\textrm{eikonal}} + \alpha\loss_{\textrm{dcp}} + \beta\loss_{\textrm{2D}},\label{eq:total_loss}
\end{equation}
where \(\loss_{\textrm{dcp}}\) and \(\loss_{\textrm{2D}}\) are the regularizations introduced in \cref{sec:dehaze_nerf},
while the photo-consistency loss, $\loss_{\textrm{color}}$, is the standard NeRF loss, and the eikonal loss, \(\loss_{\textrm{eikonal}}\), is commonly used to regularize SDF~\cite{gropp2020implicit},
\begin{align}
    \loss_{\textrm{color}}& = \frac{1}{K}\sum_{k=1}^{K}\left\|\widehat{C}_{k}(\r,\d) - C_{k}(\r,\d)\right\|_{1},\\
    \loss_{\textrm{eikonal}} &= \frac{1}{KN}\sum_{k}^{K}\sum_{n}^{N}(\|\nabla f({\mathbf{r}}_{k}(t_n))\|_2 - 1)^2,
\label{eq:loss_color}
\end{align}
where $\widehat{C}_{k}(\r,\d)$ is the pixel color. $N$ and $K$ denote the total sampling points on a ray and the total number of rays sampled per training batch.

Finally, because of the surface representation using SDF, we can optionally adopt the object masks for supervision~\cite{yariv2021volume,wang2021neus,wang2022hfs}.
Specifically, given the object mask, \(M\), the mask loss $\loss_{\textrm{mask}}$ for a sampled ray $k$ is defined as
\begin{equation}
    \loss_{\textrm{mask}} = \text{BCE}(M_k, \hat{O}_k),\label{eq:mask_loss}
\end{equation}
where $\hat{O}_k = \sum_{i=1}^{N}T_{\sigma}^{i}\alpha^{i}$ is the total weight for the clear-view surface color along the camera ray, and $\text{BCE}$ is the binary cross entropy loss.

\begin{figure*}[htbp]
\setlength{\tabcolsep}{0pt}
\renewcommand{\arraystretch}{0.8}\footnotesize
\centering\begin{tabular}{*{6}{>{\centering\arraybackslash}M{0.166\textwidth}}}
NeuS & COLMAP + NeuS &ImDehaze+NeuS& VidDehaze + NeuS & \moniker{} & Ground Truth  \\
\includegraphics[width=\linewidth]{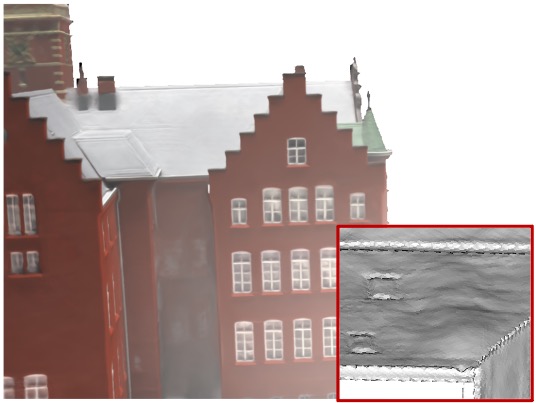}&
\includegraphics[width=\linewidth]{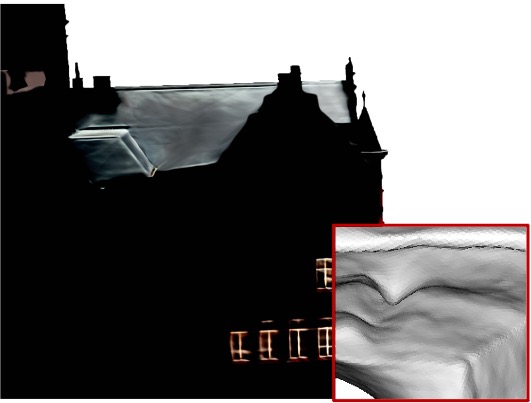} &
\includegraphics[width=\linewidth]{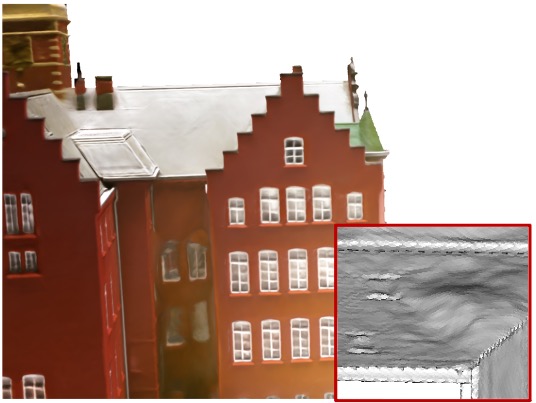}&
\includegraphics[width=\linewidth]{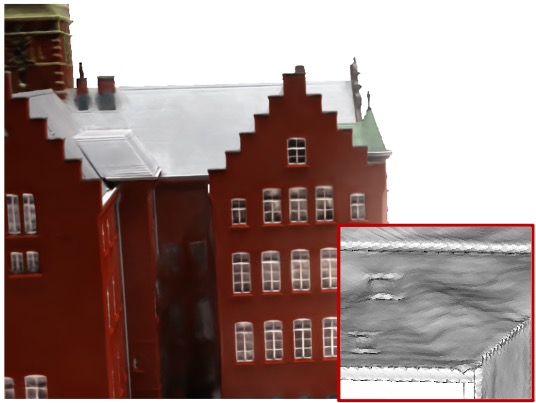}&
\includegraphics[width=\linewidth]{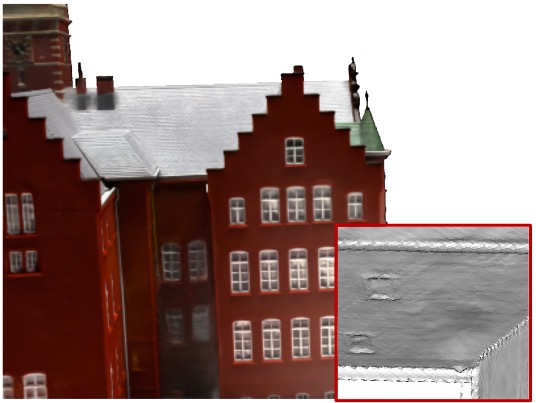}&
\hspace*{+0.4cm}\includegraphics[width=\linewidth]{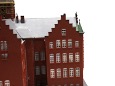}
\\
\includegraphics[width=\linewidth]{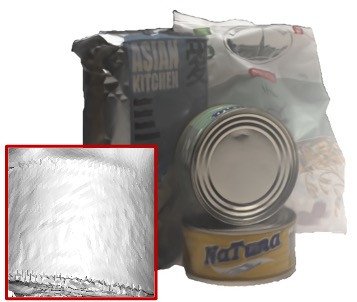}&
\includegraphics[width=\linewidth]{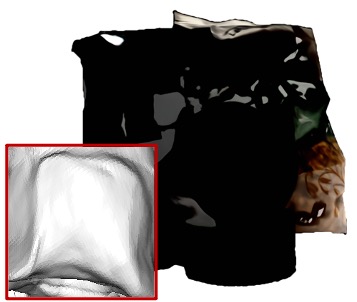} &
\includegraphics[width=\linewidth]{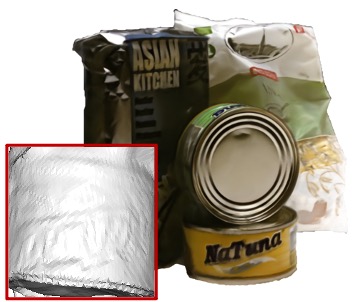}&
\includegraphics[width=\linewidth]{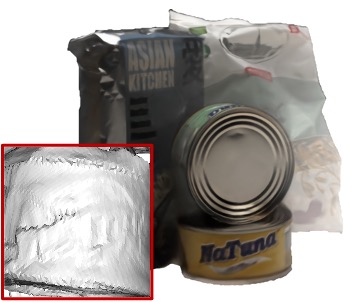}&
\includegraphics[width=\linewidth]{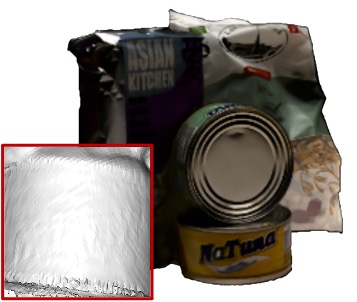}&
\hspace*{-0.2cm}\includegraphics[width=\linewidth]{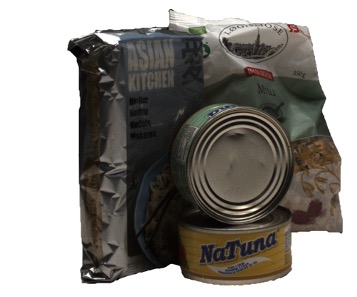}
\\
\includegraphics[width=\linewidth]{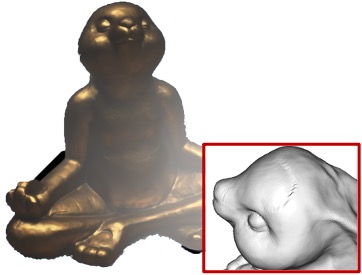}&
\includegraphics[width=\linewidth]{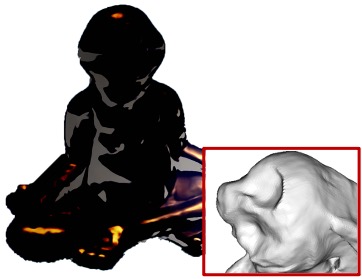}&
\includegraphics[width=\linewidth]{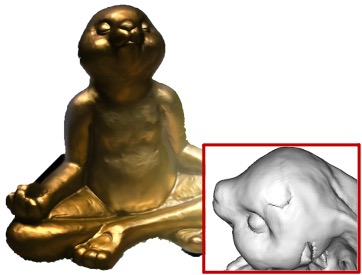}&
\includegraphics[width=\linewidth]{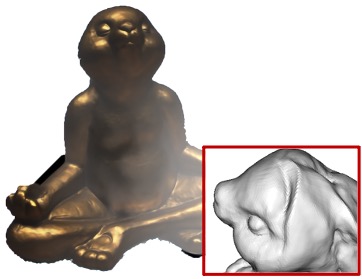}&
\includegraphics[width=\linewidth]{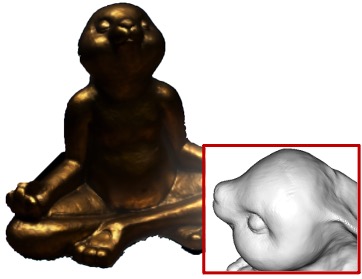}&
\hspace*{+0.6cm}\includegraphics[width=\linewidth]{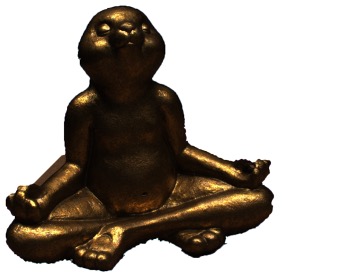}
\\

\includegraphics[width=\linewidth]{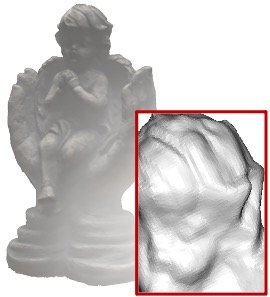}&
\includegraphics[width=\linewidth]{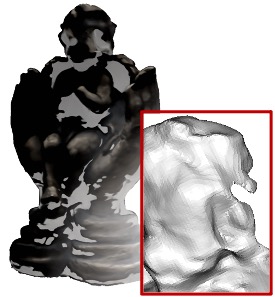}&
\includegraphics[width=\linewidth]{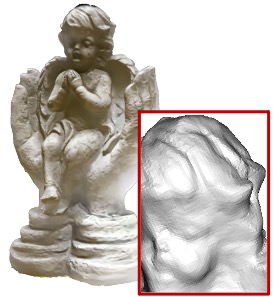}&
\includegraphics[width=\linewidth]{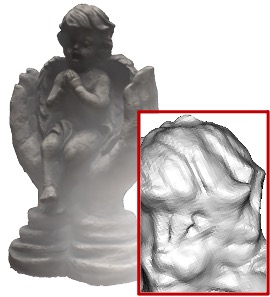}&
\includegraphics[width=\linewidth]{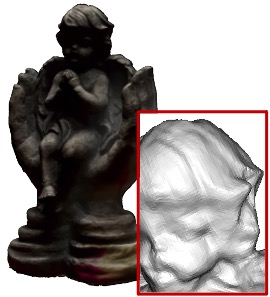}&
\hspace*{+0.9cm}\includegraphics[width=\linewidth]{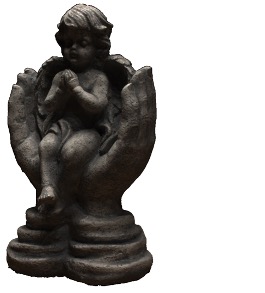}\\
\end{tabular}
\caption{\textbf{Qualitative comparison on synthetic data.} Our method successfully removes heterogeneous haze in the synthesized views, showing the best appearance fidelity compared with baseline methods. The reconstructed geometry is more accurate, less noisy, and contains more details.
}\vspace{-0.5cm}
\label{fig:dtu_qualitative}
\end{figure*}

\section{Experiments with Synthetic Scenes}
In this section, we detail our experiments using synthetic data.
Our goal is to quantifiably evaluate the contribution of each proposed component in a controlled setting.
We report our main findings here and refer the readers to the supplement for more detailed evaluations.
\subsection{Data Preparation}
We synthesize haze using 10 scenes from the DTU dataset~\cite{jensen2014large}.
% The haze synthesis follows the 2D haze generation method~\cite{li2018benchmarking} based on Koschmieder's model~\cref{eq:koschmider}.
% The depth used for haze generation is computed using the meshes reconstructed by public available NeuS models~\cite{wang2021neus}.
% We set the atmospheric light and scattering coefficient to a constant value sampled in range $\sigma_s\in[0.2, 0.6]$ and $c_{s}\in[0.5, 0.9]$.\todo{change to Gaussian blob}
The scattering coefficient is modeled using the sum of 4 scaled Gaussian blobs located inside the spatial bounding box with a standard deviation uniformly sampled from 1.0 to 3.0;
the 3-D atmospheric light is sampled from a uniform distribution in the range $[0.7, 0.9]$.
10\% of the images in each synthetic scene are held out as the test set.

\subsection{Comparisons}\label{sec:comparisons}
\paragraph{Baselines.}
We compare {\moniker} with the following baselines:
\begin{compactenum}
\item \textbf{NeuS}: train HF-NeuS~\cite{wang2022hfs} on hazy images,
\item \textbf{ImDehaze+NeuS}: train HF-NeuS on dehazed images obtained using the state-of-the-art single-image dehazing method~\cite{guo2022image},
\item \textbf{VidDehaze+NeuS}: train HF-NeuS on dehazed images obtained using the state-of-the-art video dehazing method~\cite{zhang2021learning},
\item \textbf{COLMAP+NeuS}: train HF-NeuS on dehazed images obtained  by estimating the transmission maps using the dense depth map from COLMAP~\cite{schoenberger2016sfm,schoenberger2016mvs}.
\end{compactenum}\label{lst:baseline5}
With HF-NeuS as the backbone surface model~\cite{wang2022hfs}, all approaches observe the same surface prior.
While the first baseline neglects haze entirely, baselines 2 to 4 increasingly incorporate more multiview information for haze modeling, with ours being the most 3D-aware and physically accurate, as it models the spatial-variant scattering coefficient in 3D space and optimizes the 3D geometry, surface appearance, and the haze parameters jointly.
For the last baseline, we use the method proposed by~\cite{he2010single} to estimate the global airlight from the object regions (masked by~\cite{yariv2020multiview}) in all images.
Then we use 300 pairs of feature correspondences to estimate the global scattering coefficient, where each pair computes a candidate scattering coefficient as follows
\(\frac{1}{D_{b}\left( \x_{b} \right) - D_{a}\left( \x_{a} \right)}\ln\left( \frac{I_{a}\left( \x_{a} \right) - \bar{c}_{s}}{I_{b}\left( \x_{b} \right) - \bar{c}_{s}} \right)\).
\(\left( I_{a}, D_{a} \right)\) and \(\left( I_{b}, D_{b} \right)\) are RGB images and depth maps in two views, and \(\x_{a}\) and \(\x_{b}\) are the image coordinates of a pair of matched SIFT features.
The final result is obtained after filtering out negative or invalid estimations, which may occur due to specularities and noisy depth estimation.

\noindent\textbf{Qualitative Evaluation.}
We demonstrate some examples of the dehazed results and reconstructed geometry in \cref{fig:dtu_qualitative}.
Despite having a surface prior, naïvely training HF-NeuS directly from hazy images is equivalent to averaging the scattering-induced geometry-dependent irradiance variance observed across different views and attributing it the surface color.
Consequently, the view synthesis is hazy and blurred.
For two-stage strategies, the rendered results have color distortion of various degrees, as indicated by the PSNR evaluation in \cref{tab:dtu_quantitative}, since it is difficult to accurately estimate the airlight and coefficient when the presented data does not comply with the specific assumptions or fall in the distribution of the training data.
Moreover, our method clearly reconstructs the geometry with more surface details compared to all other baselines that adopt the surface prior formulated in NeuS, since our method can dehaze different views more consistently thanks to the underlying geometry that is optimized jointly.

\noindent\textbf{Quantitative Evaluation.} We measure the image quality using peak signal-to-noise ratio (PSNR), structural similarity (SSIM), and perceptual similarity (LPIPS).
The geometry quality is measured using Chamfer Distances (CD) using the DTU standard protocol.
As shown in \cref{tab:dtu_quantitative},
\moniker~achieves the best results compared to all baselines,
with superior performance in PSNR, a metric sensitive to low-frequency color shift.
This indicates that while other methods struggle to estimate the true air light and scattering coefficient from either statistical or data priors, our method benefits from jointly optimizing these quantities along with the scene appearance and geometry.

\begin{figure}
    \centering
    \begin{subfigure}[b]{0.155\textwidth}
        \centering
        \includegraphics[width=\textwidth]{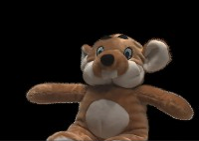}{}
        \caption{Without $\loss_{\textrm{dcp}}$}
        \label{fig:ablation_dcp_baseline}
    \end{subfigure}
    \begin{subfigure}[b]{0.155\textwidth}
        \centering
        \includegraphics[width=\textwidth]{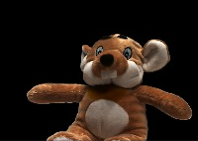}{}
        \caption{With $\loss_{\textrm{dcp}}$}
        \label{fig:ablation_dcp_ours}
    \end{subfigure}
        \begin{subfigure}[b]{0.155\textwidth}
        \centering
        \includegraphics[width=\textwidth]{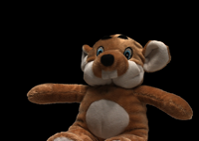}{}
        \caption{Ground Truth}
        \label{fig:ablation_dcp_gt}
    \end{subfigure}
	% \label{fig:ablation_visual}
% \end{figure}

% \begin{figure}
%     \centering
    \begin{subfigure}[b]{0.155\textwidth}
        \centering
        \includegraphics[width=\textwidth]{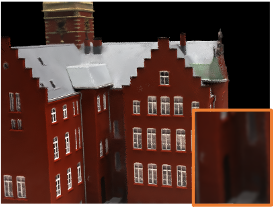}{}
        \caption{Without $\loss_{\textrm{2D}}$}
        \label{fig:ablation_cyle_base}
    \end{subfigure}
    \begin{subfigure}[b]{0.155\textwidth}
        \centering
        \includegraphics[width=\textwidth]{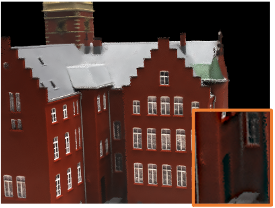}{}
        \caption{With $\loss_{\textrm{2D}}$}
        \label{fig:ablation_cyle_ours}
    \end{subfigure}
        \begin{subfigure}[b]{0.155\textwidth}
        \centering
        \includegraphics[width=\textwidth]{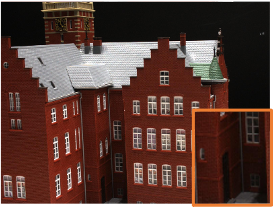}{}
        \caption{Ground Truth}
        \label{fig:ablation_cyle_gt}
    \end{subfigure}
    \label{fig:ablation_3d}\vspace{-3ex}
    \caption{\textbf{Ablation:} $\loss_{\textrm{dcp}}$ and \(\loss_{\textrm{2D}}\) lead to more accurate clear-view color photometric details.}
    \vspace{-0.5cm}
	\label{fig:ablation_visual}
\end{figure}

\begin{figure}
    \centering
    \begin{subfigure}[b]{0.23\textwidth}
        \centering
        \includegraphics[width=\textwidth]{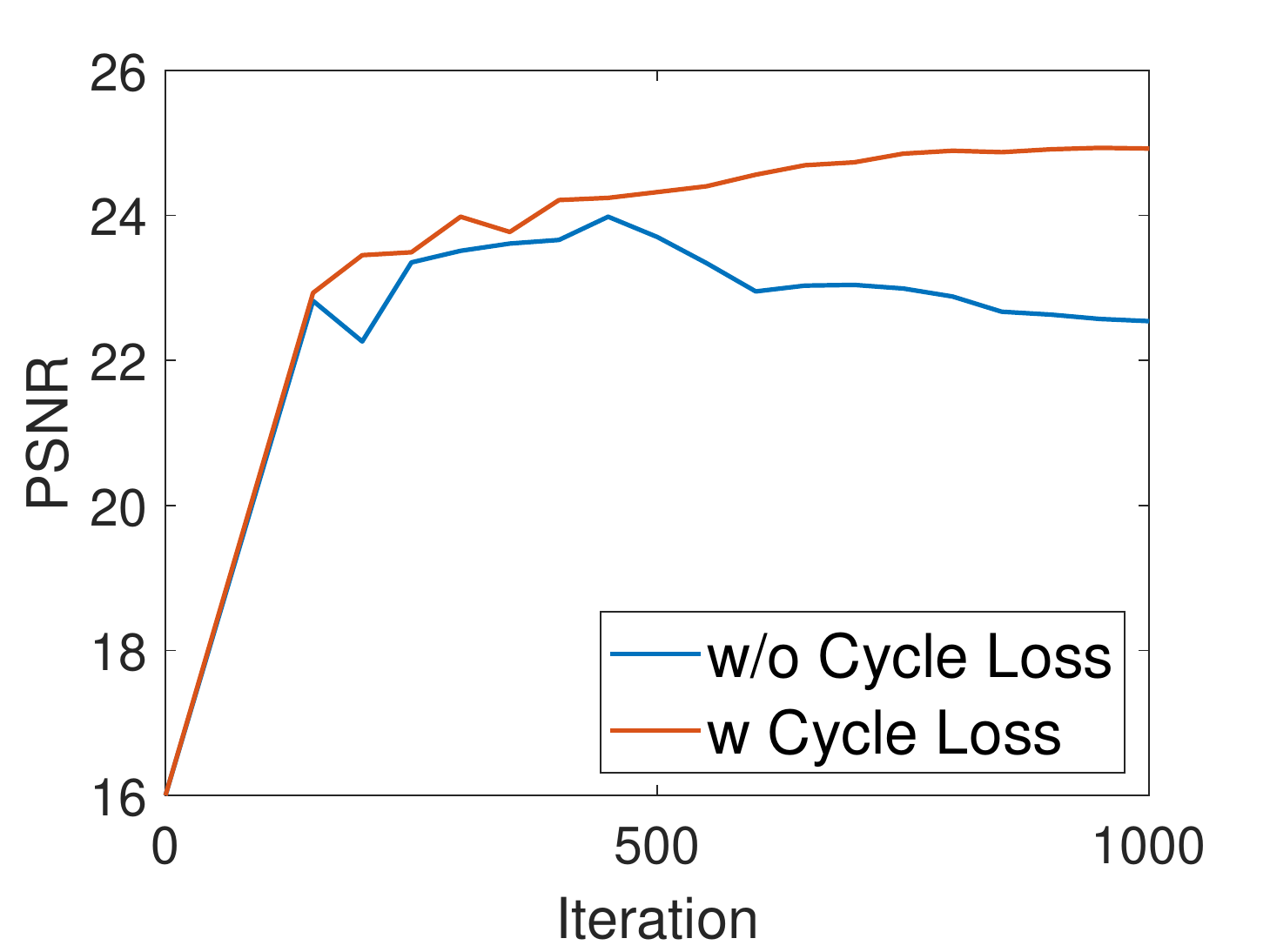}
    \end{subfigure}
        \begin{subfigure}[b]{0.23\textwidth}
        \centering
        \includegraphics[width=\textwidth]{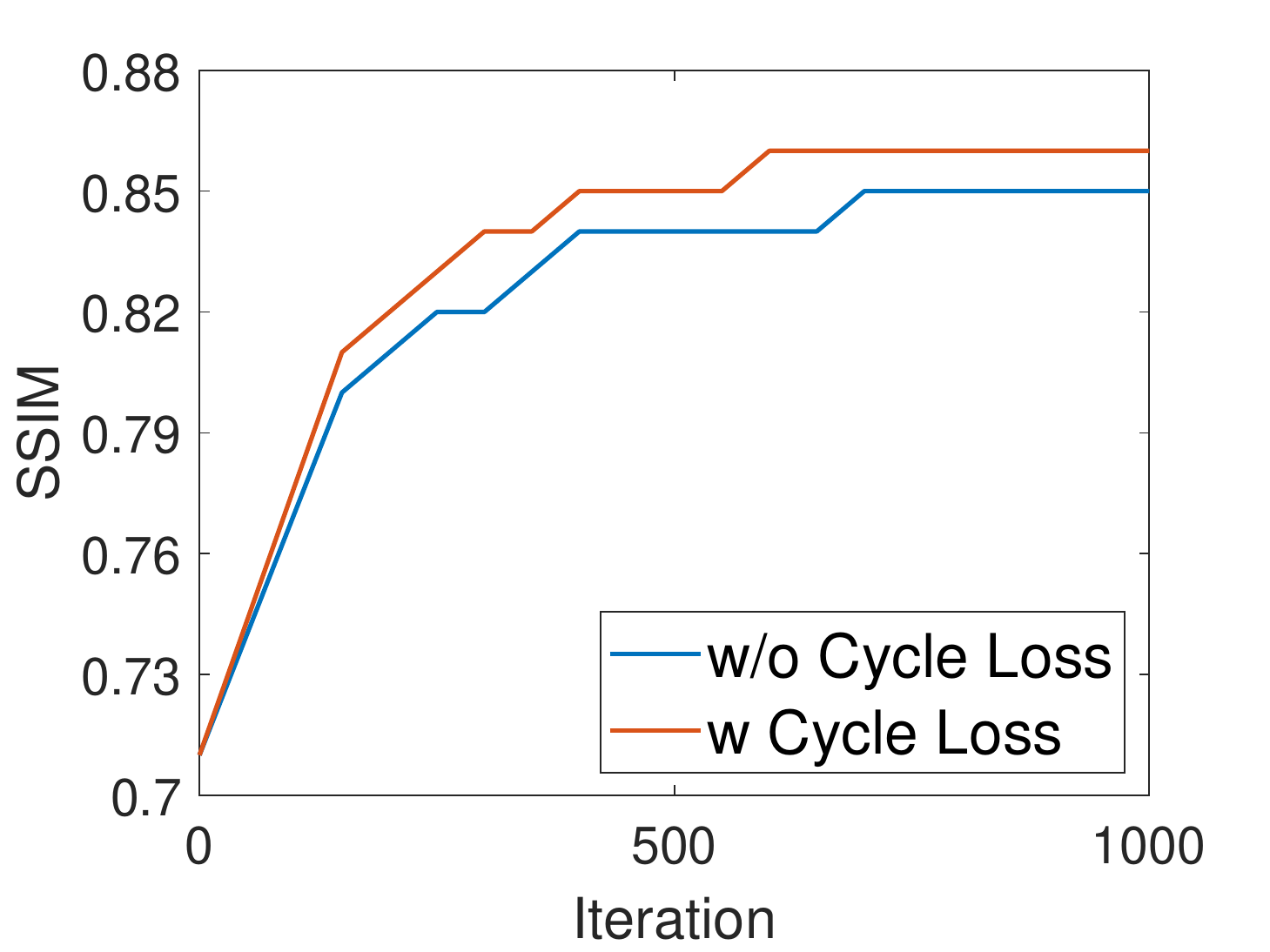}
    \end{subfigure}
 \caption{\textbf{Ablation on \(\loss_{\textrm{2D}}\).} We compare the validation results during training with and without \(\loss_{\textrm{2D}}\): With \(\loss_{\textrm{2D}}\) the image quality continuously improve.}
 \label{fig:ablation_2D_loss_curve}
\end{figure}

\begin{table}[t!]
\centering\small
\resizebox*{\linewidth}{!}{
\begin{tabular}{ccccc}\toprule
Method & PSNR ($\uparrow$) & SSIM ($\uparrow$) & LPIPS ($\downarrow$) & Chamfer ($\downarrow$)\\\midrule
        %  NeRF & 19.038	& 0.865 & 0.156	& 3.584 \\
         NeuS & 16.722 & 0.879 & 0.087 & 2.635 \\
         COLMAP + NeuS & 16.750 & 0.785 & 0.225 & 5.971 \\
         ImDehaze\cite{guo2022image} + NeuS & 17.887 & 0.899 & \underline{0.078} & \underline{2.298} \\
         VidDehaze\cite{zhang2021learning} + NeuS & \underline{18.324} & \underline{0.901} & 0.087 & 2.349 \\
         \moniker{} & \textbf{25.702} & \textbf{0.920} & \textbf{0.052} & \textbf{2.066}
\\\bottomrule
\end{tabular}
}
\caption{\textbf{Quantitative comparison using synthetic data with heterogeneous haze}. \moniker{} yields better reconstruction both in image quality and geometry accuracy.}
\label{tab:dtu_quantitative}
\vspace{-3ex}
\end{table}

\subsection{Ablation Study}\label{sec:ablation}
We conduct ablation studies for the optimization regularizers on 3 scenes and report their average in \cref{tab:ablation_1}.

As \cref{tab:ablation_1} shows, both \(\loss_{\textrm{2D}}\) and \(\loss_{\textrm{dcp}}\) contribute positively to the image quality and surface reconstruction accuracy.
The \(\loss_{\textrm{dcp}}\) has a prominent effect in reducing global color shift as indicated by the PSNR value.
Similarly, from the visual comparison in \cref{fig:ablation_dcp_baseline,fig:ablation_dcp_ours,fig:ablation_dcp_gt}, we can observe that by adopting \(\loss_{\textrm{dcp}}\), the residual haze can be suppressed effectively.
Moreover, as shown in~\cref{fig:ablation_cyle_base,sub@fig:ablation_cyle_ours,sub@fig:ablation_cyle_gt}, the results optimized with \(\loss_{\textrm{2D}}\) show more accurate structure (see orange bounding boxes).
In addition, we plot the evolution of validation PSNR and SSIM in \cref{fig:ablation_2D_loss_curve}.
%One can observe that
\(\loss_{\textrm{2D}}\) improves the convergence behavior and yields better validation results.

\begin{table}[t!]
\centering
\resizebox*{\linewidth}{!}{
\begin{tabular}{ccccc}\toprule
Method & PSNR ($\uparrow$) & SSIM ($\uparrow$) & LPIPS ($\downarrow$) & Chamfer ($\downarrow$)\\\midrule
w/o $\loss_{\textrm{2D}} + \loss_{\textrm{dcp}}$ & 18.21 & 0.88 & 0.09 & 2.47 \\
w $\loss_{\textrm{2D}}$ & 18.53 & 0.89 & 0.09 & 2.44  \\
w $\loss_{\textrm{dcp}}$ & 24.02 & 0.90 & 0.06 & 2.44 \\
\moniker{} & \textbf{25.23} & \textbf{0.91} & \textbf{0.05} & \textbf{2.41} \\
\bottomrule
\end{tabular}
}
\caption{\textbf{Ablations on regularization}. Each of the proposed two regularizations can individually improve the image and geometry reconstruction, and the best quality is achieved with both. The results are averaged from 10 test scenes from the DTU dataset.}
\vspace{-0.2cm}
\label{tab:ablation_1}
\end{table}

\begin{figure*}[t!]
\setlength{\tabcolsep}{0pt}
\renewcommand{\arraystretch}{0.75}\footnotesize
\centering\begin{tabular}{*{6}{>{\centering\arraybackslash}M{0.165\textwidth}}}
 NeuS & COLMAP + NeuS& ImDehaze + NeuS & VidDehaze + NeuS & \moniker{} & Ground Truth \\
\includegraphics[width=0.98\linewidth, clip, trim={2cm 2cm 0cm 5cm}]{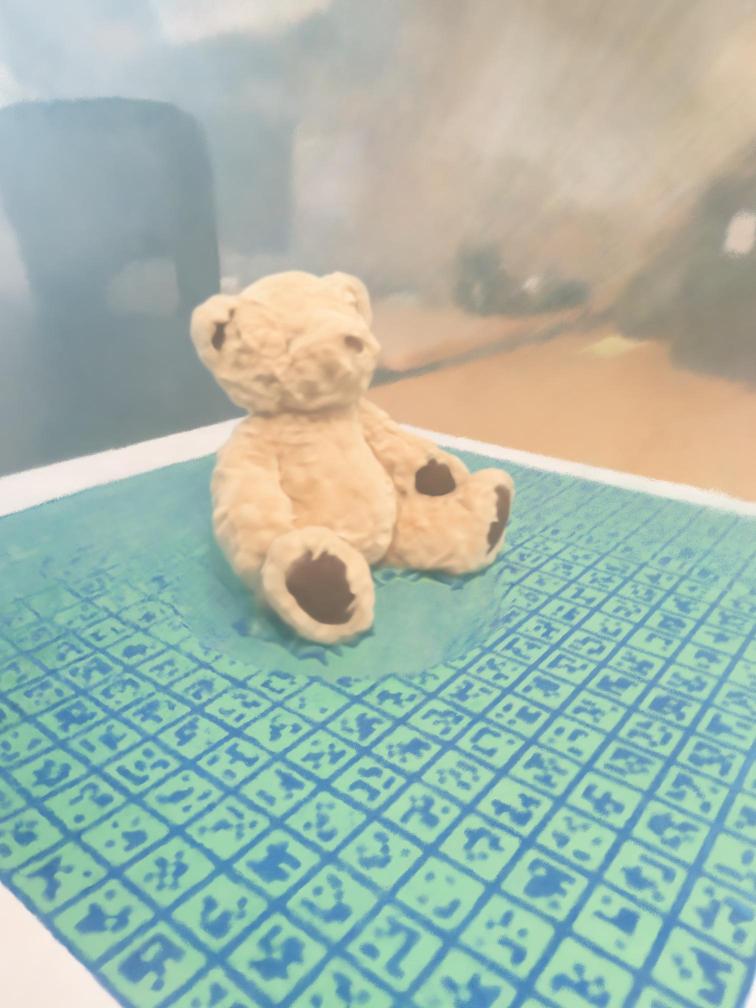} &
\includegraphics[width=0.98\linewidth, clip, trim={2cm 2cm 0cm 5cm}]{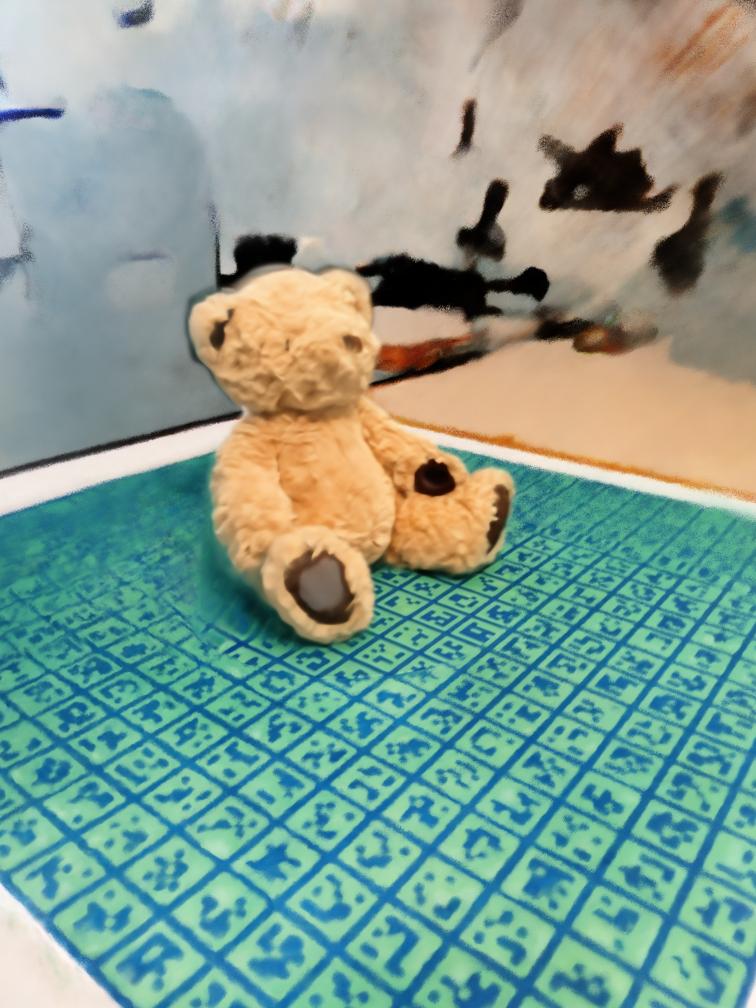} &
\includegraphics[width=0.98\linewidth, clip, trim={2cm 2cm 0cm 5cm}]{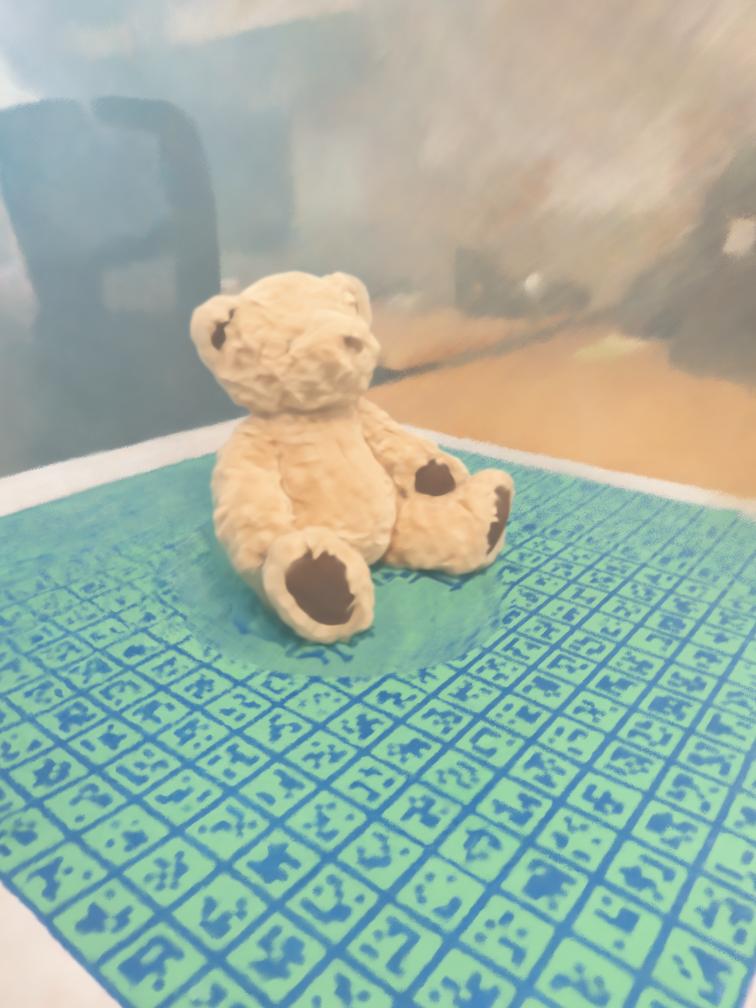} &
\includegraphics[width=0.98\linewidth, clip, trim={2cm 2cm 0cm 5cm}]{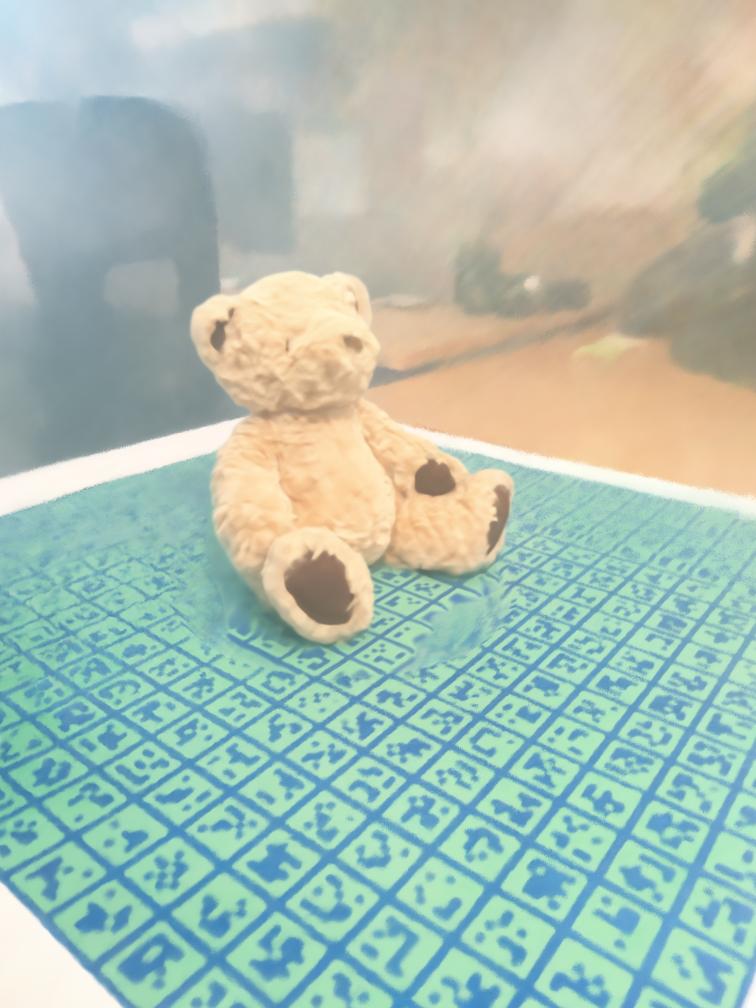} &
\includegraphics[width=0.98\linewidth, clip, trim={0.5cm 0.5cm 0cm 1.25cm}]{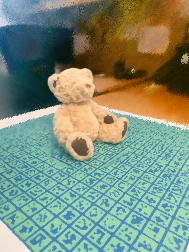} &
\includegraphics[width=0.98\linewidth, clip, trim={2cm 2cm 0cm 5cm}]{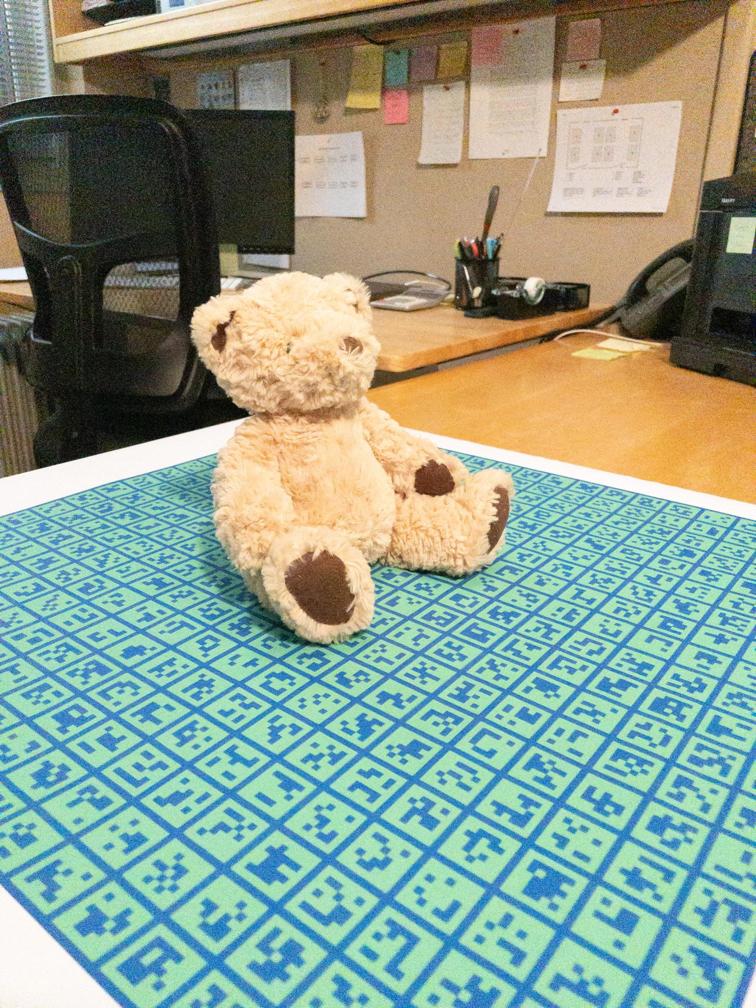}
\\
\includegraphics[width=0.98\linewidth, clip, trim={0cm 1cm 0.5cm 0cm}]{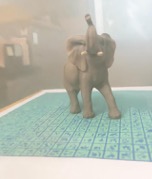} &
\includegraphics[width=0.98\linewidth, clip, trim={0cm 1cm 0.5cm 0cm}]{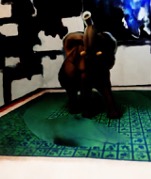} &
\includegraphics[width=0.98\linewidth, clip, trim={0cm 1cm 0.5cm 0cm}]{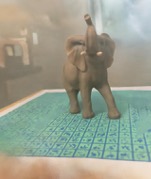} &
\includegraphics[width=0.98\linewidth, clip, trim={0cm 1cm 0.5cm 0cm}]{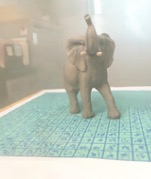} &
\includegraphics[width=0.98\linewidth, clip, trim={0cm 1cm 0.5cm 0cm}]{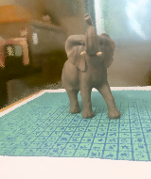} &
\includegraphics[width=0.98\linewidth, clip, trim={0cm 1cm 0.5cm 0cm}]{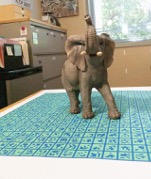}
\\
\includegraphics[width=0.98\linewidth, clip, trim={0cm 2.5cm 0cm 0cm}]{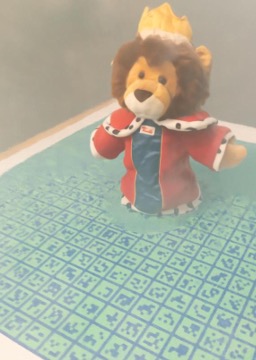} &
\includegraphics[width=0.98\linewidth, clip, trim={0cm 2.5cm 0cm 0cm}]{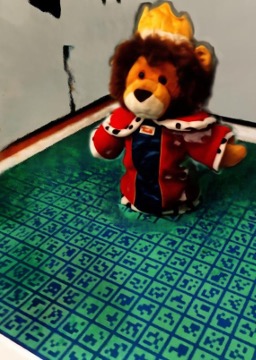} &
\includegraphics[width=0.98\linewidth, clip, trim={0cm 2.5cm 0cm 0cm}]{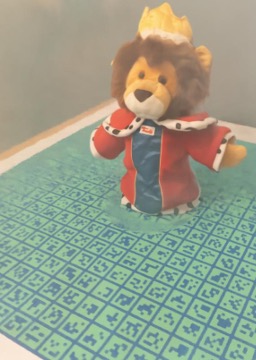} &
\includegraphics[width=0.98\linewidth, clip, trim={0cm 2.5cm 0cm 0cm}]{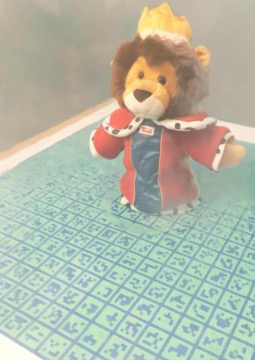} &
\includegraphics[width=0.98\linewidth, clip, trim={0cm 2.5cm 0cm 0cm}]{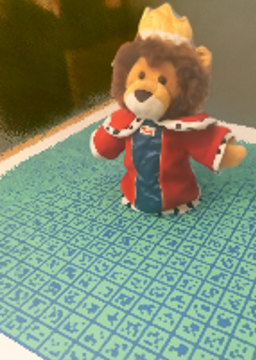} &
\includegraphics[width=0.98\linewidth, clip, trim={0cm 2.5cm 0cm 0cm}]{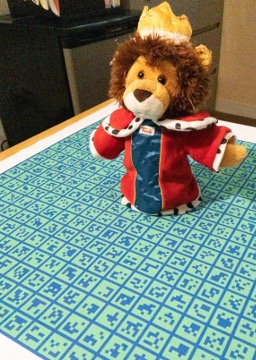}
\\
\end{tabular}
\caption{\textbf{Qualitative comparison on captured data}. Our method recovers the haze-free scenes with more image details and colors closer to the ground truth.}
\vspace{-3ex}
\label{fig:real_qualitative}
\end{figure*}

\begin{table}[t!]
\resizebox*{\linewidth}{!}{
\begin{tabular}{cccc}\toprule
Method & PSNR ($\uparrow$) & SSIM ($\uparrow$) & LPIPS ($\downarrow$) \\\midrule
% NeRF & 11.98&0.48 &0.37 \\
NeuS & 12.60& 0.48& 0.38\\
COLMAP + NeuS &9.27 & 0.38&0.47 \\
ImDehaze + NeuS  & 13.35 & 0.49& 0.36\\
VidDehaze + NeuS  &\underline{14.56} & \underline{0.50}& \underline{0.34} \\
\moniker{} & \textbf{17.47} & \textbf{0.68} & \textbf{0.16}\\\bottomrule
\end{tabular}
}
\caption{\textbf{Quantitative evaluation on experimentally captured data} averaged over 3 scenes. Our method outperforms other methods by a large margin.}
\label{tab:real_quantitative}
\vspace{-0.5cm}
\end{table}

\section{Experimentally Captured Results}
\subsection{Data Collection}
We captured three indoor hazy scenes using two professional haze machines and an iPhone 12 Pro.
The scenes contain 38, 82, and 68 hazy images and 79, 47, and 47 clear images for testing respectively.
To ensure temporal consistency (dynamic haze is out of scope for this paper), we capture the data after the haze has settled and appears steady.
After the capturing, we merge hazy images and clear images under the same scene and adopt COLMAP to register camera positions, which yields a consistent registration for the hazy and clear view for ease of evaluation.
We scale and translate the object such that it lies inside the unit sphere and roughly at the center of the world coordinates.
More details on the data capture are described in the supplement.

\subsection{Implementation details.}
Several changes are applied to the model to facilitate the optimization for captured data.
First, we remove the mask loss (~\cref{eq:mask_loss}).
Second, similar to NeuS~\cite{wang2021neus}, the region outside uses the parameterization of NeRF++\cite{zhang2020nerf++}, for which we also applied the haze attenuation following \cref{eq:C_surface,eq:alpha,eq:C_haze}.

\subsection{Comparison}
For evaluation, we adopt the same baselines described in \cref{sec:comparisons}.
The 2D photometric quality is evaluated using the same metrics, and geometry reconstruction quality is omitted due to the lack of ground truth.

\paragraph{Quantitative Evaluation.}
The results are reported in \cref{tab:real_quantitative}.
One can observe that our method outperforms other baselines in all metrics, demonstrating the robustness of the proposed method in the real-world scenario.

\paragraph{Qualitative Evaluation.}
As shown in \cref{fig:real_qualitative},
vanilla NeuS includes haze in the synthesized views.
On the other hand, adopting 2D image dehazing or video dehazing methods as pre-processing also cannot generate desired clear results since these learning-based models have limited generalization ability in real-world data.
Although using COLMAP as pre-processing can suppress residual haze, the dehazed results tend to have the color distortion problem and the limited structure (see the background at the $3^{rd}$ column of \cref{fig:real_qualitative}).
Our method achieves significantly better visual quality compared to other baselines.

\section{Discussion}
\label{sec:discussion}
In this paper, we address challenges to apply NeRFs in hazy scenes.
Our method jointly learns the 3D structure and the plausible clear-view appearance of the scene, as well as the haze properties (airlight and scattering coefficient) from hazy observations.
Our key technical contributions are:
1. incorporating the scattering phenomena in NeRF's rendering equation;
2. deploying physically inspired inductive biases and haze-specific regularizers to disambiguate haze and clear-view components.
The proposed techniques take a significant step towards adopting NeRF-related approaches in real-life scenarios where adverse weather conditions are detrimental to 3D reconstruction.

\paragraph{Relation to the Koschmieder Law.}
Koschmieder model is a 2D image formation model that describes the relationship between the clear-view and the hazy images.
Assuming homogeneous scattering, our formulation can be simplified to Koschmieder's model, which reduces to supervising with \cref{eq:koschmieder} and \(\loss_{\textrm{dcp}}\).
Our \(\loss_{\textrm{2D}}\) exploits this relation to improve the convergence as shown in \cref{sec:ablation}.
In \cref{tab:koschmieder}, we examine this relation by comparing this simplified method (denoted as Koschm) with a variant of \moniker{} that uses a learnable scalar instead of the proposed MLP to model the scattering coefficient (denoted as Scalar).
For both synthetic and real data, the Scalar and Koschm solutions are comparable, which validates the relation to Koschmieder law; our full model shows a considerable advantage in both scenarios thanks to its more general formulation.
\begin{table}[htbp]
\scriptsize
\scalebox{0.92}{
\begin{tabular}{@{\hspace{-0.02em}}c*{7}
{P{0.07\linewidth}}}
\toprule
\multirow{2}{*}{Method} & \multicolumn{4}{c}{Synthetic Data (Scene24)} & \multicolumn{3}{c}{Real Data (``Elephant'' Scene)} \\\cmidrule(lr){2-5} \cmidrule(lr){6-8}
 & PSNR & SSIM & LPIPS & Chamfer & PSNR & SSIM & LPIPS \\\midrule
Koschm & 18.62 & 0.79 & 0.13 & 1.44 & 16.84 & 0.69 & 0.18 \\
Scalar & 18.50 & 0.79 & 0.13 & 1.46 & 16.80 & 0.69 & 0.18 \\
\moniker{} & \textbf{22.78} & \textbf{0.82} & \textbf{0.09} & \textbf{1.22} & \textbf{17.87} & \textbf{0.73} & \textbf{0.15} \\
\bottomrule
\end{tabular}}
    \caption{\textbf{Relation to the 2D Koschmieder Law.} When using a scalar to model the scattering coefficient (Scalar), our method can simplify to supervising with the 2D Koschmieder law (Koschm). Our full model using an MLP to model the spatially varying scattering coefficient has a clear advantage for heterogeneous haze.}
    \label{tab:koschmieder}
    \vspace{-0.5cm}
\end{table}

\paragraph{Limitations and future work.}
While we have demonstrated the effectiveness of our method to capture data, certain scenarios are not addressed in the scope of this paper and are research directions for future work.
These include severe haze scenes, where camera registration fails due to the lack of discriminative image features, and dynamic haze, where the haze distribution change not only spatially but also temporally.
Nonetheless, we believe that our method is an important step towards anticipated application scenarios, such as autonomous driving or underwater imaging.
The principle of incorporating physics into the neural rendering is also applicable to other ill-posed low-level vision tasks, such as image denoising, image brightening, and super-resolution.
\ificcvfinal
\section*{Acknowledgement}
This project was in part supported by Samsung, Stanford HAI, a PECASE from the ARO, SNF Postdoc.mobility fellowship, and Google PhD Fellowship. We thank to National Center for High-performance Computing (NCHC) for providing computational and storage resources.
\fi

% \clearpage
%%%%%%%%% REFERENCES

{\small
\bibliographystyle{ieee_fullname}
\bibliography{egbib}
}
%%%%%%%%% Appendix
\appendix

\begin{center}
\textbf{\large Supplemental Materials}
\end{center}
%%%%%%%%%% Merge with supplemental materials %%%%%%%%%%
%%%%%%%%%% Prefix a "S" to all equations, figures, tables and reset the counter %%%%%%%%%%
\renewcommand{\theequation}{S.\arabic{equation}}
\renewcommand\thefigure{S.\arabic{figure}}
\renewcommand\thetable{S.\arabic{table}}
\setcounter{equation}{0}
\setcounter{figure}{0}
\setcounter{table}{0}

\begin{table}[tbhp]
\setlength{\columnsep}{0pt}
    \centering\scalebox{0.75}{
    \begin{tabular}{cc*{4}{c}}\toprule
Data & Method & PSNR $\uparrow$ & SSIM $\uparrow$ & LPIPS $\downarrow$ & Chamfer $\downarrow$\\\midrule
\multirowcell{5}{Scan\\24}
         % & NeRF & 17.37& 0.72& 0.30 & 3.72 \\
         & NeuS & 13.93 & 0.72 & 0.19 & \underline{1.51} \\
         & COLMAP+NeuS & 11.82 & 0.41 & 0.54 & 6.22 \\
         & ImDehaze\cite{guo2022image}+NeuS & \underline{16.83} & \underline{0.80} & \underline{0.18} & 1.60  \\
         & VidDehaze\cite{zhang2021learning}+NeuS & 16.73 & 0.77 & 0.19 & 1.59 \\
         & \moniker{} & \textbf{22.78} &\textbf{0.82} &\textbf{0.09} &\textbf{1.22} \\\midrule
\multirowcell{5}{Scan\\37}
         % & NeRF & 21.05& 0.86& 0.15 & 3.70 \\
         & NeuS & 20.64 & 0.88 & \underline{0.06} & \underline{1.44} \\
         & COLMAP+NeuS & 13.34 & 0.75 & 0.23 & 3.04\\
         & ImDehaze\cite{guo2022image}+NeuS & \underline{21.55} & \underline{0.90} & 0.06 & 1.90\\
         & VidDehaze\cite{zhang2021learning}+NeuS & 20.90 & 0.88 & 0.08 & 2.16\\
         & \moniker{} & \textbf{21.61} & \textbf{0.90} & \textbf{0.06} & \textbf{1.43} \\\midrule
% \multirow{5}{*}{Scan 40}& NeRF & 19.77& 0.77& 0.42 & 2.84 \\
%          & NeuS & \underline{19.86} & 0.79&\underline{0.10} & \textbf{1.21} \\
%          & COLMAP+NeuS & 11.05& 0.38& 0.68& 3.54 \\
%          & ImDehaze\cite{guo2022image}+NeuS & 21.21 &\underline{0.82} & 0.12 &1.72 \\
%          & VidDehaze\cite{zhang2021learning}+NeuS & 19.59 & 0.81 & 0.12&\underline{1.71} \\
%          & \moniker{} & \textbf{24.86} &\textbf{0.87} &\textbf{0.09} &2.45 \\\midrule
\multirowcell{5}{Scan\\83}
         % & NeRF & 19.54 & 0.93& 0.06& 8.45\\
         & NeuS & 20.25 & 0.94 & 0.05 & 7.03 \\
         & COLMAP+NeuS & 17.23 & 0.92 & 0.08 & 7.10 \\
         & ImDehaze\cite{guo2022image}+NeuS & \underline{22.89} & 0.94 & 0.05 & \underline{7.01} \\
         & VidDehaze\cite{zhang2021learning}+NeuS & 21.12 & \underline{0.95} & \underline{0.05} & 7.03 \\
         & \moniker{} & \textbf{28.46} & \textbf{0.95}& \textbf{0.03}& \textbf{6.91}\\\midrule
\multirowcell{5}{Scan\\97}
         % & NeRF & 21.27 & 0.89& 0.14& 2.68 \\
         & NeuS & 17.32 & 0.91 & 0.07 & 2.08\\
         & COLMAP+NeuS & 16.93 & 0.77 & 0.25 & 2.96 \\
         & ImDehaze\cite{guo2022image}+NeuS & 18.18 & \underline{0.93} & 0.07& \underline{2.02}\\
         & VidDehaze\cite{zhang2021learning}+NeuS & \underline{18.61} & \underline{0.93} & \underline{0.07} & 2.05\\
         & \moniker{} & \textbf{24.41} & \textbf{0.92}&\textbf{0.05} &\textbf{1.89} \\\midrule
\multirowcell{5}{Scan\\105}
         % & NeRF &19.76 &0.92 &0.11 &4.08 \\
         & NeuS & \underline{18.98} & 0.93& 0.07 &2.65 \\
         & COLMAP+NeuS & 15.65 & 0.77& 0.33& 3.78 \\
         & ImDehaze\cite{guo2022image}+NeuS & 18.84 & \underline{0.94} & \underline{0.06} & \textbf{2.64}\\
         & VidDehaze\cite{zhang2021learning}+NeuS & 17.74 & 0.92& 0.09 & 2.95 \\
         & \moniker{} & \textbf{26.18} & \textbf{0.95}& \textbf{0.04}&\underline{2.65} \\\midrule
\multirowcell{5}{Scan\\106}
         % & NeRF & 17.22 &0.84 &0.15 &3.23 \\
         & NeuS & 14.57& 0.84 & 0.11 & 4.24 \\
         & COLMAP+NeuS & 17.03 & 0.78 & 0.22 & 2.29 \\
         & ImDehaze\cite{guo2022image}+NeuS & 15.45 & 0.86 & \underline{0.09} & \underline{1.34} \\
         & VidDehaze\cite{zhang2021learning}+NeuS & \underline{17.20} & \underline{0.89} & 0.11 & 1.67 \\
         & \moniker{}& \textbf{28.41} & \textbf{0.92} & \textbf{0.07} & \textbf{1.23}\\\midrule

\multirowcell{5}{Scan\\110}
         % & NeRF & 20.77&0.87 &0.11 &3.36 \\
         & NeuS & 15.21 & 0.86 & 0.13 & \underline{1.90} \\
         & COLMAP+NeuS & 19.01 & 0.84 & 0.20 & 2.88 \\
         & ImDehaze\cite{guo2022image}+NeuS & \underline{17.53} & \underline{0.89} & \underline{0.10} & 2.30 \\
         & VidDehaze\cite{zhang2021learning}+NeuS & 15.69 & 0.86 & 0.13 & 2.11\\
         & \moniker{}&  \textbf{23.28} & \textbf{0.93} & \textbf{0.06} & \textbf{1.76} \\
         \midrule

\multirowcell{5}{Scan\\114}
         & NeuS & 16.88 & 0.90 & 0.07 & 0.86 \\
         & COLMAP+NeuS & 15.60 & 0.82 & 0.17 & 1.19 \\
         & ImDehaze\cite{guo2022image}+NeuS & 17.39 & 0.92 & \textbf{0.06} & \textbf{0.80} \\
         & VidDehaze\cite{zhang2021learning}+NeuS & \underline{19.96} & \textbf{0.93} & \underline{0.06} & 0.87\\
         & \moniker{}&  \textbf{22.73} & \underline{0.91} & \underline{0.06} & \underline{0.84} \\\midrule

\multirowcell{5}{Scan\\118}
         & NeuS & 14.47 & 0.86 & 0.09 & 2.64 \\
         & COLMAP+NeuS & \underline{18.28} & 0.82 & 0.17 & 1.90 \\
         & ImDehaze\cite{guo2022image}+NeuS & 15.14 & 0.88 & \underline{0.07} & 1.66 \\
         & VidDehaze\cite{zhang2021learning}+NeuS & 17.57 & \underline{0.91} & 0.08 & \underline{1.53}\\
         & \moniker{}&  \textbf{28.28} & \textbf{0.92} & \textbf{0.05} & \textbf{1.25} \\\midrule

\multirowcell{5}{Scan\\122}
         & NeuS & 15.35 & 0.90 & 0.07 & 1.23 \\
         & COLMAP+NeuS & 19.83 & 0.85 & 0.17 & 1.57 \\
         & ImDehaze\cite{guo2022image}+NeuS & 15.95 & 0.92 & 0.06 & 1.35 \\
         & VidDehaze\cite{zhang2021learning}+NeuS & \underline{18.05} & \underline{0.92} & \underline{0.08} & \underline{1.22}\\
         & \moniker{}&  \textbf{28.55} & \textbf{0.94} & \textbf{0.06} & \textbf{0.99} \\\midrule

\multirowcell{5}{Average}
         &NeuS & 16.72 & 0.88 & 0.09 & 2.64 \\
         &COLMAP+NeuS & 16.75 & 0.79 & 0.23 & 5.97 \\
         &ImDehaze\cite{guo2022image}+NeuS & 17.89 & 0.90 & \underline{0.08} &\underline{2.30} \\
         &VidDehaze\cite{zhang2021learning}+NeuS & \underline{18.32} & \underline{0.90} & 0.09 & 2.35 \\
         &\moniker{} & \textbf{25.70} & \textbf{0.92} & \textbf{0.05} & \textbf{2.07}\\
    \bottomrule
    \end{tabular}}
    \caption{Quantitative evaluation for synthetic haze dataset.}
    \label{tab:dtu_quantitative}
\end{table}

\section{Data generation}
\subsection{Synthetic Data}
\begin{figure}[htbp]
    \centering
\includegraphics[width=0.49\linewidth]{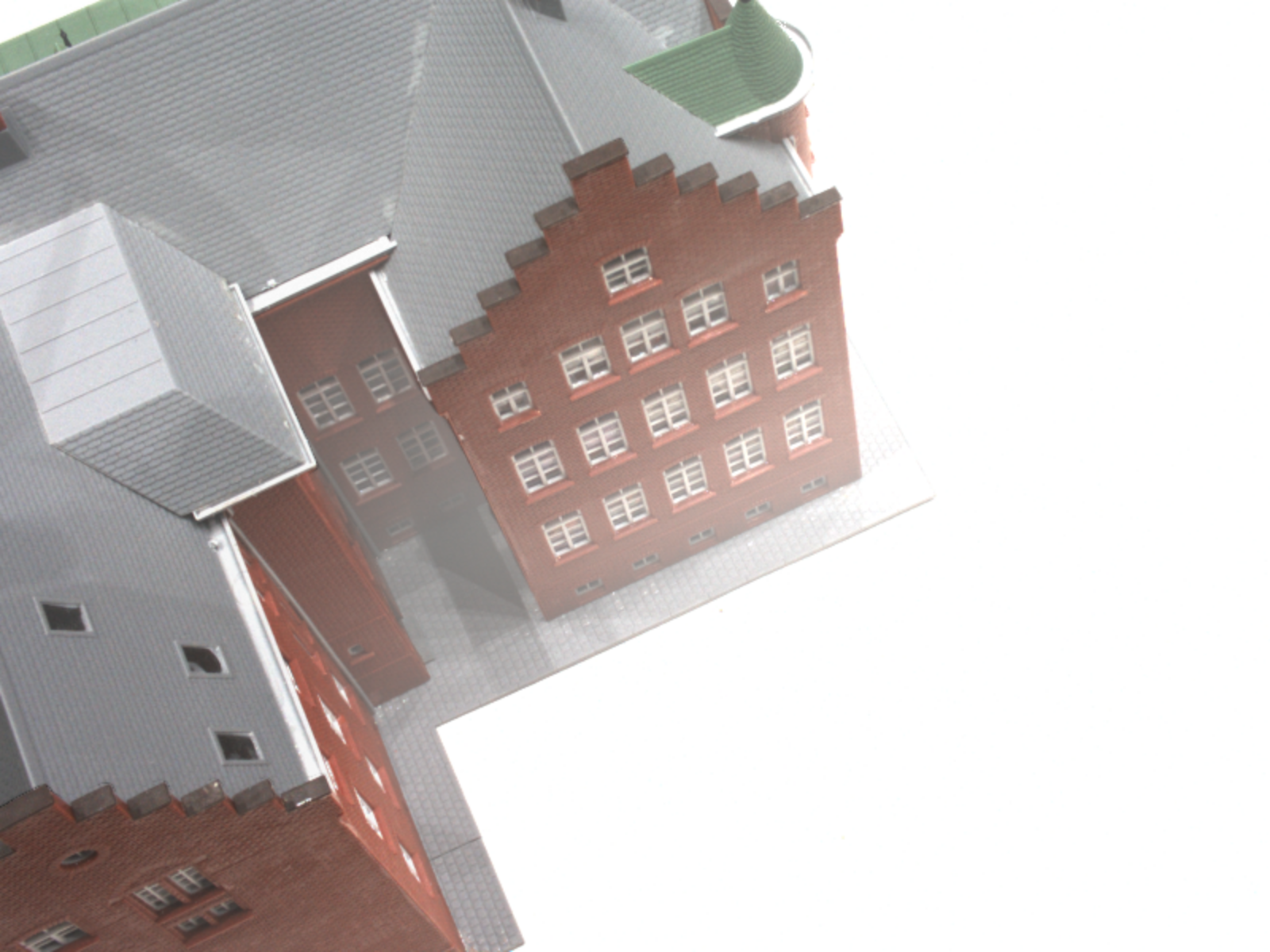}%
\includegraphics[width=0.49\linewidth]{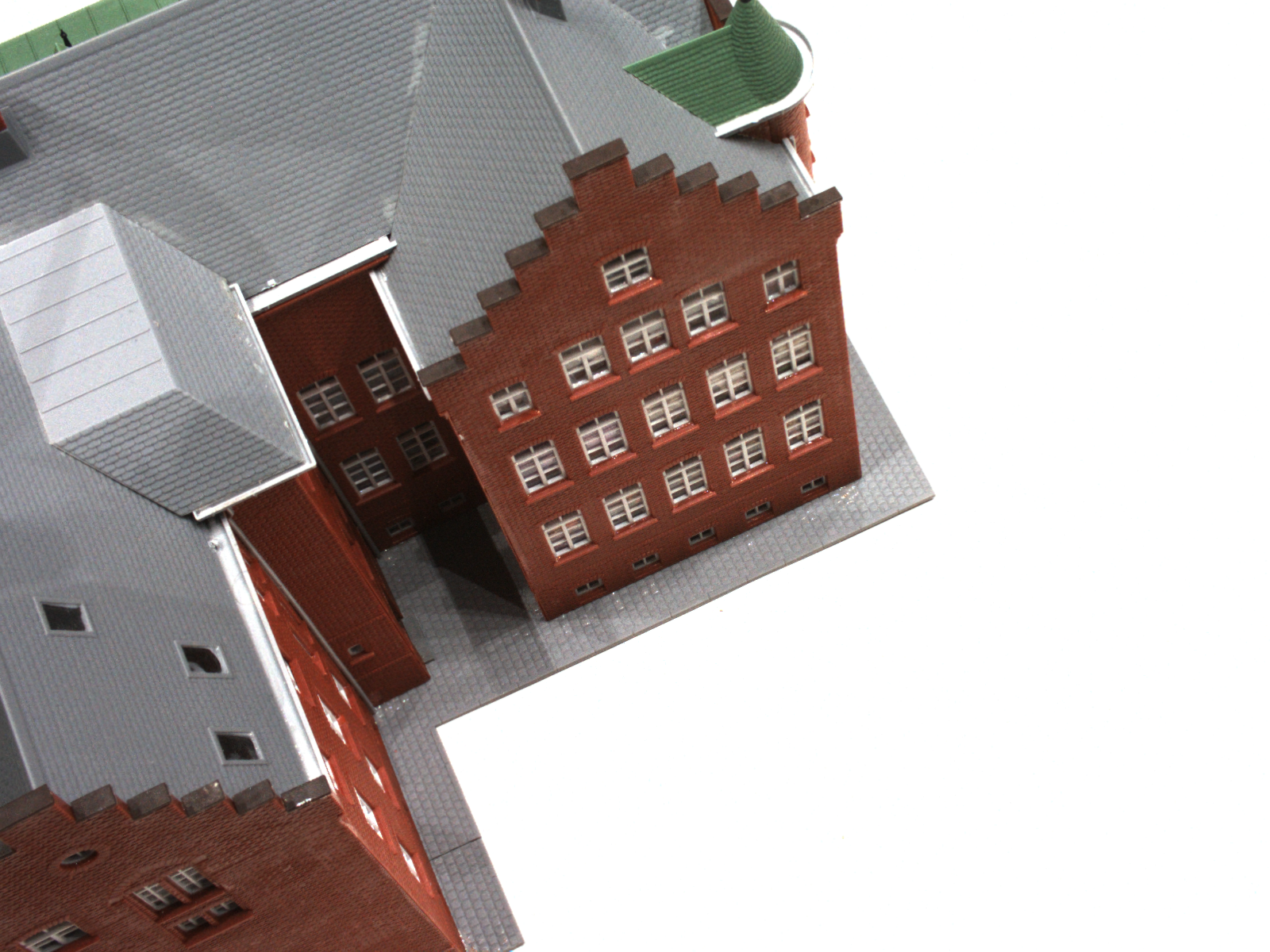}
    \caption{\textbf{An example of the synthetic data.} Right: original input image. Left: synthetic hazy input with heterogeneous scattering.}
    \label{fig:synthetic_data}
\end{figure}

An example of our synthesized results are shown in \cref{fig:synthetic_data}.
The scattering coefficient is modeled using the sum of 4 scaled Gaussian blobs located inside the spatial bounding box with a standard deviation uniformly sampled from 1.0 to 3.0;
the 3-D atmospheric light is sampled from a uniform distribution in the range $[0.7, 0.9]$.

\subsection{Real Data Collection}
We use two professional haze machines to generate a dense vapor. These haze generators adopt cast or platen-type aluminum heat exchangers, which make evaporation of the water-based haze liquid. We capture our scenes in a hermetic chamber.
First, we captured clear reference images from different view directions.
Then, we employed 2 minutes for both haze machines and waited for another 2 minutes until the haze was temporally stable.
We then captured hazy images from different viewpoints.
An iPhone 12 was used as our camera.
The camera's settings including exposure, iso, and focal length were the same for all captures.

After the capturing, we merge hazy images and clear images under the same scene and run COLMAP~\cite{schoenberger2016sfm,schoenberger2016mvs} to register camera positions.
% Then, we adopt hazy images to train our network and clear images to test our network.

\section{Evaluation}

\paragraph{Quantitative Evaluation on Synthetic Dataset}
We present the complete quantitative results on DTU dataset~\cite{jensen2014large} in~\cref{tab:dtu_quantitative}. One can see that the proposed \moniker{} outperforms other baselines in most of the scenes in terms of image quality and geometric reconstruction.

% \paragraph{Dehazing under Different Haze Densities}
% We present the comparison of our method in different haze densities and atmospheric light intensities with other baselines in \cref{tab:haze_density}. Our method is better than other baselines in different haze densities and atmospheric light intensities.
% Also, one can observe that the performance of dehazing degrades when the haze density increases (i.e., $\beta$). On the other hand, when the atmospheric light intensity increases (i.e., $A$), the performance of dehazing becomes better.

\begin{table}
\center\small
\begin{tabular}{ccccc}\toprule
Surface Prior & PSNR$\uparrow$ & SSIM$\uparrow$ & LPIPS$\downarrow$ & Chamfer$\downarrow$\\\midrule
None & 21.331 & 0.901 & 0.071 & 4.013 \\
via NeuS & \textbf{26.039} & \textbf{0.921} & \textbf{0.056} & \textbf{2.863}
\\\bottomrule
\end{tabular}
\caption{The effect of NeuS surface prior evaluated on 5 test scenes.}
\label{tab:ablation_backbone}
\end{table}

\paragraph{Additional Ablation}
Here, we present an additional ablation to demonstrate the effect of using the NeuS surface prior.
To this end, we replace the NeuS backbone with a NeRF~\cite{mildenhall2020nerf} backbone and compare the image quality and geometry reconstruction performance.
As shown in~\cref{tab:ablation_backbone}, by adopting a surface prior, implemented via the surface-based parameterization of the volume density~NeuS, the model can converge to a better solution in terms of haze-surface disambiguation leading to significantly better image and geometry reconstruction quality.

\section{Implementation Detail}
\paragraph{Network architecture}
We adopt a similar network architecture as HF-NeuS~\cite{wang2022hfs}. It contains two MLPs to encode surface SDF and surface color respectively. The SDF MLP consists of 8 hidden layers with hidden size of 256. A skip connection is applied from the input to the fourth hidden layer.
The color MLP contains 4 hidden layers with size of 256. It takes the spatial location \textbf{p}, the view direction \textbf{d}, the normal vector of SDF, and a 256-dimensional feature vector from the SDF MLP.
Positional encoding is applied to spatial location and view direction with 6 and 4 frequencies, respectively.

For the spatially variant scattering coefficient, we adopt band-limited coordinate networks~\cite{lindell2022bacon} using 3 hidden layers, where each layer has 32 channels and the frequency parameter set to 10.

\paragraph{Training and inference details}
We train the proposed \moniker{} by using the Adam optimizer~\cite{kingma2014adam}. We set the learning rate linearly from 0 to $5\times10^{-4}$ by warmed up strategy in the first 5k iterations. Then, we reduce the learning rate by the cosine decay schedule to the minimum learning rate $2.5\times10^{-5}$. We train our model for 24 hours (for the ‘w/ mask’ setting) and 26 hours (for the ‘w/o mask’ setting) on a single Nvidia Tesla V100 GPU with batch size 512, respectively. For the first 300k iterations, we set the scattering coefficients and atmospheric light as zero and train the model based on the vanilla HF-NeuS. In this stage, we disable the Koschmieder Consistency and dark channel losses. After 300k iterations, the scattering coefficients and atmospheric light are introduced and optimized with the Koschmieder Consistency and dark channel losses. We train our network for 800k and 50k iterations for synthetic and captured data, respectively. The scale factors $\lambda$, $\alpha$, and $\beta$ in the loss function are set to 0.1, 5000, and 0.01 for synthetic data and 0.1, 1.0, and 0.01 for the captured dataset, respectively. The scale factor of mask loss is set to 0.1. For the inference stage, it takes 200 seconds to render an image in resolution of $1600\times1200$.

\section{Deriving \moniker{}}

\subsection{Radiative Transfer Equation}
Radiative transfer equation (RTE)~\cite{chandrasekhar2013radiative,van1999multiple} describes the behaviour of light in a medium that absorbs, scatters and emits radiation.
Assuming, a ray \(\r\left( t \right) = \mathbf{o} + t\d\) hits a surface point at \(\r\left( t_{0} \right)\), the incident radiance at the near image plane \(t_{n}\) can be divided into three parts~\cite{pharr2016physically}:

\begin{align}\begin{split}
C(\r, \d)=&
\underbrace{\int_{t_{n}}^{t_{0}}\epsilon\left(\r\left( t\right),\d\right)T_{\sigma_{t}}\left( t\right)dt}_{\textrm{emission}}\\
&+\underbrace{\int_{t_{n}}^{t_{0}}c_{\textrm{s}}\left( \r\left( t \right), \d \right)\sigma_{s}\left(\r\left( t \right)\right)T_{\sigma_{t}}\left( t \right)dt}_{\textrm{in-scattering}}\\
&+\underbrace{C_e\left(\r\left( t_{0} \right),\d\right)T_{\sigma_{t}}\left( t_{0}\right)}_{\textrm{surface reflection}},\label{eq:RTE}
\end{split}
\end{align}
where \(\epsilon\) is the emission, \(C_{e}\) is the outgoing radiance at the surface intersection, \(c_{\textrm{s}}\left(\r\left( t \right), \d \right)\) is the in-scattered light and \(\sigma_{s}\) is the scattering coefficient.
In particular the transmittance here is computed from the attenuation coefficient \(\sigma_{t}\), \ie,
\(T_{\sigma_{t}}\left( t\right)=\exp\left( -\int_{t_{n}}^{t}\sigma_{t}(t')dt' \right)\),
where \(\sigma_{t}=\sigma_{a} + \sigma_{s}\) including the absorption and out-scattering effect.

Based on this equation, we then connect it to the neural radiance field (NeRF) and hazy image formation model.

\subsection{From RTE to NeRF}
\label{sec:4.2}
For the volume rendering equation adopted in NeRF, there are two assumptions: (i) no scattering and (ii) only absorption and emission are considered. The volume rendering equation can be derived by the RTE equation by dropping the in-scattering term and combining the surface reflection term into the emission term. It can be presented as:
\cite{max1995optical,wang2021neus,yariv2021volume}
% \yifan{I think it would be better if you directly replace \(\sigma_{t}\) with \(\sigma_{a}\) here, since you mentioned we drop scattering.}
\begin{align}\begin{split}
C(\r, \d)=&
\underbrace{\int_{t_{n}}^{t_{0}}\epsilon\left(\r\left( t\right),\d\right)T_{\sigma_{a}}\left( t\right)dt}_{\textrm{emission}}
\\&+\underbrace{C_e\left(\r\left( t_{0} \right),\d\right)T_{\sigma_{a}}\left( t_{0}\right)}_{\textrm{surface reflection}}.
\label{eq:RTE_haze_formation}
\end{split}
\end{align}

Since NeRF treats each point as an emission point, the radiance of the surface can be considered as emission at $t_0$. Thus, the surface reflection in \cref{eq:RTE_haze_formation} becomes
\begin{equation}
\resizebox{0.9\linewidth}{!}{
\(
    C_e\left(\r\left( t_{0} \right),\d\right)T_{\sigma_{a}}\left( t_{0}\right) = \int_{t_n}^{t_0}\epsilon_{o}\left(\r\left( t_{0} \right),\d\right)\delta(t- t_{0})T_{\sigma_{a}}\left( t\right)dt.
    \)

}
\label{eq:replace_surface}\end{equation}
Therefore \cref{eq:RTE_haze_formation} can be written as:
\begin{align}
\begin{split}
C(\r, \d)&=
% \int_{t_{n}}^{t_{0}}\left(\epsilon\left(\r\left( t\right),\d\right)+\epsilon_{o}\left(\r\left( t_{0} \right),\d\right)\right)\delta(t -t_{0})T_{\sigma_{a}}\left( t\right)dt\\
\int_{t_{n}}^{t_{0}}\epsilon_{\textrm{combined}}\left(\r\left( t\right),\d\right)T_{\sigma_{a}}\left( t\right)dt,
\label{eq:NeRF_RTE}
\end{split}\end{align}
where \(\epsilon_{\textrm{combined}}=\epsilon\left(\r\left( t\right),\d\right)+\epsilon_{o}\left(\r\left( t_{0} \right),\d\right)\delta(t-t_{0})\).

By further setting $\epsilon_{\textrm{combined}}\left(\r\left( t\right),\d\right)=c(\r(t), \mathbf{d})\sigma(\r(t))$ and \(\sigma_{a} = \sigma\), we can see \cref{eq:NeRF_RTE} is the same as famous NeRF rendering equation below:

\begin{gather}
\begin{split}
{C}(\r, \d)=\int_{t_n}^{t_f}c(\r(t), \mathbf{d})\sigma(\r(t))T(t) \ dt.
\label{eq:NeRF}
\end{split}
\end{gather}

% We can see \cref{eq:NeRF} can be derived from \cref{eq:NeRF_RTE} by setting $\epsilon_{\textrm{combined}}\left(\r\left( t\right),\d\right)=c(\r(t), \mathbf{d})\sigma(\r(t))$ and denoting \(\sigma_{a} = \sigma\).
In practice, absorption coefficient $\sigma_{a}(\r(t))$ describes the probability of a photon being absorbed at position $\r\left( t \right)$, while $\sigma(\r(t))$ (called volume density in NeRF) illustrates the probability of a ray terminating at location $\r\left( t \right)$.
That is, $\sigma_{a}$ is the same meaning as $\sigma$ physically since NeRF only considers the absorption.
Moreover, generally, we can assume that when the ray hits the object, the attenuation term (accumulative transmittance) may become zero. That is, $T(t)\cong0$ when $t>t_{0}$. Thus, the integral from $t_n$ to $t_f$ is approximately equal to that from $t_n$ to $t_0$.

\subsection{From RTE to Koschmieder Model}
For the conventional haze formation model, there are three assumptions\cite{middleton1957vision,narasimhan2004models}: (i) no emission, (ii) only consider single scattering, and (iii) only consider scattering since absorption in atmospheric particles is relatively small. Therefore, \cref{eq:RTE} simplifies to
\begin{align}
C(\r, \d)=&
\underbrace{\int_{t_{n}}^{t_{0}}c_{\textrm{s}}\left( \r\left( t \right), \d \right)\sigma_{s}\left(\r\left( t \right)\right)T_{\sigma_{s}}\left( t \right)dt}_{\textrm{in-scattering}}\nonumber\\
&+\underbrace{C_e\left(\r\left( t_{0} \right),\d\right)T_{\sigma_{s}}\left( t_{0}\right)}_{\textrm{surface reflection}}.\label{eq:RTE_Haze}
\end{align}

Assuming the scattering coefficient and atmospheric light are isotropic and homogeneous, \ie, \(\sigma_{s}\left( \r\left( t \right) \right) \equiv \bar{\sigma}_{s}\) and \(c_{s}\left( \r\left( t \right), \d\right)\equiv\bar{c}_{s}\), setting $t_{n}=0$ and \(C_e\left(\r\left( t_{0} \right),\d\right)=C_{\textrm{clear}}\), we can rewrite \cref{eq:RTE_Haze} to the Koschmieder model~\cite{israel1959koschmieders}, namely
\begin{equation}
 \resizebox{1\hsize}{!}{
 $
\begin{split}
C(\r, \d)&={\bar{c}_{\textrm{s}}\int_{0}^{t_0}\bar{\sigma}_{s}\underbrace{e^{-\int_{0}^{t}\bar{\sigma}_{s}dt^{'}}}_{T_{\bar{\sigma}_{s}}\left( t \right)}dt}
+
{C_e\left(\r\left( t_{0} \right),\d\right)\underbrace{e^{-\int_{0}^{t}\bar{\sigma}_{s}dt^{'}}}_{T_{\bar{\sigma}_{s}}\left( t \right)}\left( t_{0}\right)}\\
&=\bar{c}_{s}(1-\exp(-\bar{\sigma}_{s} t_{0})) + C_{\textrm{clear}}\exp(-\bar{\sigma}_{s} t_{0}).
\end{split}\label{eq:koschmieder}$}
\end{equation}

\subsection{3D Haze Formation in \moniker{}}
In this part, we bring \cref{eq:RTE_Haze} to a similar form as NeRF so that we can learn the geometry and color through samples in the 3D space.

First, we bring back volume density $\sigma$, \ie, replace \(\sigma_{s}\) with \(\sigma_{t} =\sigma_{s} + \sigma\):
\begin{equation}
\begin{split}
C(\r, \d)=
\int_{t_{n}}^{t_{0}}c_{\textrm{s}}\left( \r\left( t \right), \d \right)\sigma_{s}\left(\r\left( t \right)\right)T_{\sigma_{t}}\left( t \right)dt\\
+
C_e\left(\r\left( t_{0} \right),\d\right)T_{\sigma_{t}}\left( t_{0}\right),
\end{split}
\end{equation}
which corresponds to Eq. (3) in the main paper.
Using the same trick in \cref{eq:replace_surface}, we can combine the surface reflection term into the integral and obtain the rendering equation used in \moniker{} (Eq. (4) in the main paper):
\begin{align}
C(\r, \d)=&
\int_{t_{n}}^{t_{0}}c_{\textrm{s}}\left( \r\left( t \right), \d \right)\sigma_{s}\left(\r\left( t \right)\right)T_{\sigma_{t}}\left( t \right)dt\nonumber\\
&+
\int_{t_n}^{t_0}\underbrace{\epsilon_{\textrm{surface}}\left(\r\left( t_{0} \right),\d\right)}_{\epsilon_{o}\left(\r\left( t_{0} \right),\d\right)\delta(t-t_{0})}T_{\sigma_{t}}\left( t\right)dt\nonumber\\
=&{\int_{t_{n}}^{t_{0}}c_{\textrm{s}}\left( \r\left( t \right), \d \right)\sigma_{s}\left(\r\left( t \right)\right)T_{\sigma_{t}}\left( t \right)dt}\nonumber\\
&+
{\int_{t_{n}}^{t_{0}}\underbrace{c\left( \r\left( t \right), \d \right)\sigma\left(\r\left( t \right)\right)}_{\epsilon_{\textrm{surface}}\left(\r\left( t_{0} \right),\d\right)}T_{\sigma_{t}}\left( t \right)dt}.
\end{align}
The first term (denoted as $C_{\text{haze}}$) captures the radiance caused by haze scattering. The second term (denoted as $C_{\textrm{surface}}$) is equivalent to NeRF for a scene with only solid surfaces.

\clearpage

\end{document}